%% file: main.tex
\documentclass[11pt]{article}

\usepackage[preprint]{acl}

\usepackage{times}
\usepackage{latexsym}
\usepackage[T1]{fontenc}
\usepackage[utf8]{inputenc}
\usepackage{microtype}
\usepackage{inconsolata}
\usepackage{graphicx}
\usepackage{xurl}

\usepackage{amsmath}
\usepackage{amssymb}
\usepackage{amsfonts}
\usepackage{bbm}
\usepackage{bm}
\usepackage{booktabs}
\usepackage{colortbl}
\usepackage{float}
\usepackage{nicefrac}
\usepackage{enumitem}
\usepackage{multirow}
\usepackage[ruled,noend,linesnumbered]{algorithm2e}
\usepackage{listings}
\usepackage{upquote}
\usepackage{subcaption}
\usepackage{cleveref}
\usepackage[most]{tcolorbox}

\definecolor{promptbg}{RGB}{233,240,255}
\newtcolorbox{mppaprompt}[1]{
  colback=promptbg!100!white,
  sharp corners=south,
  boxrule=0.25mm,
  fonttitle=\bfseries,
  title={#1},
  width=\textwidth,
  enhanced,
  drop shadow
}
\newcommand{\promptfield}[1]{\textcolor{blue}{\texttt{\{#1\}}}}

\input{math_commands.tex}

\newcommand{\methodname}{MPPA}

\title{Advancing the Pareto Frontier of Test-Time Scaling with Multi-Path Plan Aggregation}

\author{
  \textbf{Siheng Xiong}\textsuperscript{1},
  \textbf{Ali Payani}\textsuperscript{2},
  \textbf{Faramarz Fekri}\textsuperscript{1} \\
  \textsuperscript{1}Georgia Institute of Technology \quad
  \textsuperscript{2}Cisco Research \\
  \texttt{sxiong45@gatech.edu} \quad
  \texttt{apayani@cisco.com} \quad
  \texttt{fekri@ece.gatech.edu}
}

\begin{document}
\maketitle

\begin{abstract}
Test-time scaling seeks to improve reasoning through additional inference-time computation, commonly using solution sampling with majority voting to select the final answer.
However, repeated sampling can reproduce correlated errors, limiting the accuracy gains from additional computation.
We introduce Multi-Path Plan Aggregation (\methodname{}), which learns planning and answer selection using separate LoRA adapters on a shared frozen backbone.
The planner jointly generates $K$ short plans, each guiding a separate solution generated by the frozen executor; the aggregator then scores the resulting solutions and selects the final answer.
Joint planning encourages complementary approaches suited to the executor's capabilities, while learned selection uses reasoning evidence to identify correct candidates.
Alternating optimization improves planning and aggregation through supervised finetuning and reinforcement learning.
Across five mathematics and science benchmarks and three backbones, \methodname{} improves accuracy--token trade-offs compared with training-free and training-based baselines.
With three plans, it achieves higher average accuracy than eight directly sampled solutions with majority voting or the same learned aggregator, while reducing output tokens by 47--54\%.
Evaluation with task-provided subtasks further demonstrates higher final-task accuracy than direct sampling with majority voting or learned aggregation at either the task or subtask level.
\end{abstract}

\section{Introduction}
\label{sec:introduction}
\input{sections/introduction.tex}

\section{Effective Scaling Depends on Both Generation and Selection}
\label{sec:motivation}
\input{sections/motivation.tex}

\section{Multi-Path Plan Aggregation}
\label{sec:method}
\input{sections/method.tex}

\section{Experiments}
\label{sec:experiments}
\input{sections/experiments.tex}

\section{Related Work}
\label{sec:related_work}
\input{sections/related_work.tex}

\section{Conclusion}
\label{sec:conclusion}
\input{sections/conclusion.tex}

\section*{Limitations}
\input{sections/limitations.tex}

\section*{Acknowledgments}
This work is supported in part by DARPA SciFy program, Award No.HR001125C0302, and CISCO Systems, Inc.

\bibliography{references}

\appendix
\input{sections/appendix.tex}

\end{document}

%% file: math_commands.tex
\usepackage{amsmath,amsfonts,bm}

\def\eqref#1{equation~\ref{#1}}

\def\1{\bm{1}}

\DeclareMathAlphabet{\mathsfit}{\encodingdefault}{\sfdefault}{m}{sl}
\SetMathAlphabet{\mathsfit}{bold}{\encodingdefault}{\sfdefault}{bx}{n}



%% file: sections/introduction.tex
\begin{figure}[t]
\centering
\includegraphics[width=0.9\linewidth]{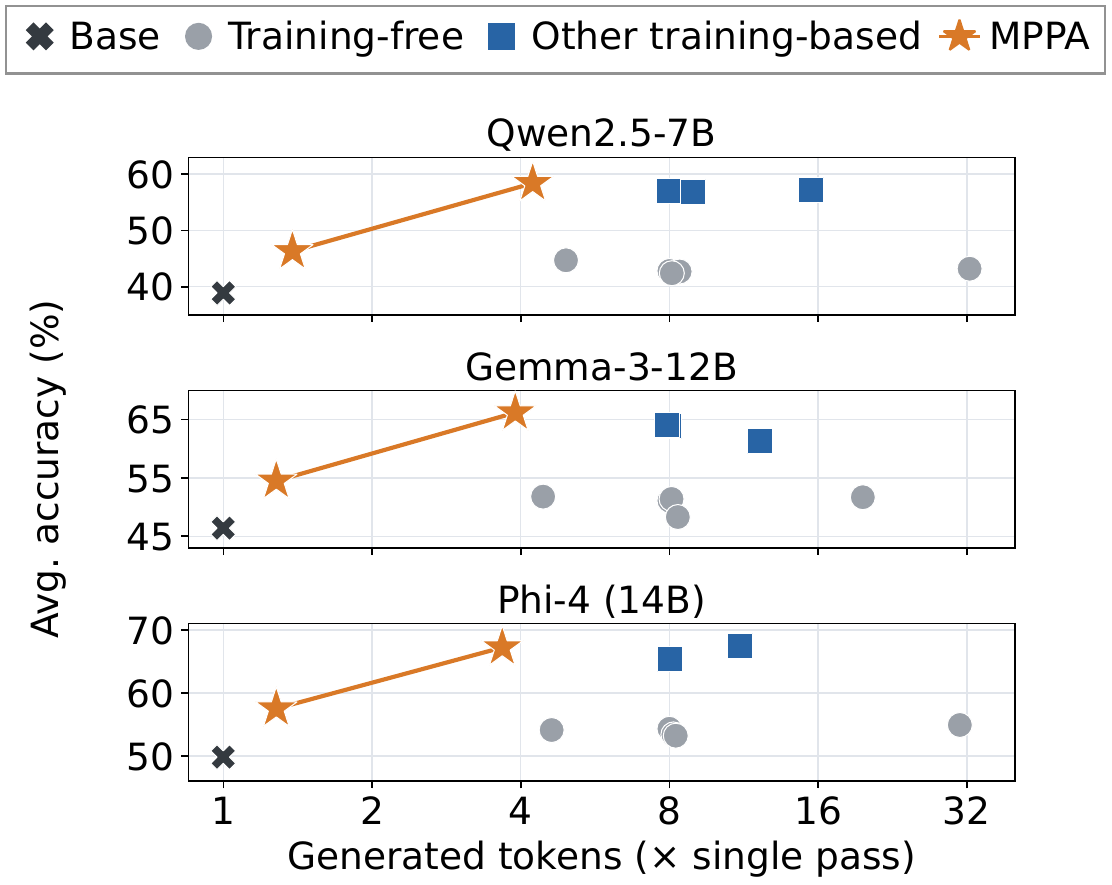}
\caption{Average accuracy versus generated-token cost across five benchmarks. Training-free includes maj@8, PS+ maj@8, L2M maj@8, ESC, and USC; Other training-based includes Base, SFT, and GRPO with agg@8, where agg denotes selection using the same learned aggregator. MPPA connects the one- and three-plan settings. Costs are normalized to a single Base model solve; See details in \Cref{tab:main_results}.}
\label{fig:main_results_overview}
\end{figure}

Test-time scaling improves reasoning performance through additional inference-time computation~\citep{openai2024reasoning,guo2025deepseek}, commonly by sampling multiple solutions and selecting an answer through majority voting~\citep{wang2023selfconsistency, wu2024inferencescaling}. Its benefit, however, can be limited by correlated errors: repeated sampling may revisit similar reasoning strategies and reproduce the same systematic mistakes, increasing computation without improving answer accuracy~\citep{brown2024monkeys}. Diverse plans offer a way to direct solution attempts toward different approaches, creating complementary opportunities for success. Such diversity must also account for the executor's capabilities. A plan proposed by a stronger model may require knowledge, reasoning steps, or computations that a weaker executor cannot reliably carry out, and thus may fail to improve end-to-end performance. These considerations motivate learning both how to propose useful, complementary plans and how to aggregate the resulting solutions. We study this problem with a frozen executor, retaining the original model as the solver while training planning and aggregation modules around its existing capabilities.

We introduce Multi-Path Plan Aggregation (\methodname{}), which learns to generate and aggregate multiple reasoning paths around a frozen executor. Given a problem and a requested number of plans $K$, the planner produces $K$ short reasoning plans in a single call, each outlining an approach to solving the problem. Joint generation allows subsequent plans to account for approaches already proposed, encouraging complementary strategies. The executor solves the original problem once under each plan, producing $K$ complete candidate solutions. The aggregator estimates the correctness of each candidate given the problem and uses these scores to select the final answer. This makes answer selection sensitive to the reasoning evidence within candidates, allowing a reliable minority answer to be favored over a frequent but incorrect one. We implement the planner and aggregator as separate LoRA adapters on a shared frozen backbone, enabling specialization through a small set of trainable parameters while reusing the underlying model weights. Both adapters are disabled during execution.

We initialize the planner through supervised learning from plan sets refined using execution feedback, with training examples requesting different numbers of plans. The aggregator is initialized through supervised training on the correctness of candidates produced by this planner and the frozen executor. We then improve the two adapters through alternating optimization. With the aggregator fixed, reinforcement learning updates the planner using a reward that combines final-answer correctness after aggregation with the presence of a correct candidate among the execution paths. With the planner fixed, supervised learning updates the aggregator using candidates collected from the current planning policy. This jointly improves candidate generation and answer selection.

\input{sections/motivation_figures.tex}

\input{sections/framework_figure.tex}

We evaluate \methodname{} on five mathematics and science benchmarks using Qwen2.5-7B-Instruct, Gemma-3-12B-it, and Phi-4. The results demonstrate improved accuracy--token trade-offs compared with training-free and training-based baselines (\Cref{fig:main_results_overview}). Across all three backbones, three plans achieve higher average accuracy than eight directly sampled solutions with majority voting or the same learned aggregator, using roughly half as many output tokens. This shows the value of learning executor-compatible plans and effective selection at a smaller candidate budget. Evaluation with task-provided subtasks further demonstrates higher final-task accuracy than direct sampling with majority voting or learned aggregation at either the task or subtask level.

Our main contributions are:
\begin{itemize}
    \item We introduce \methodname{}, which jointly generates multiple plans, executes each with a frozen executor, and selects among the resulting solutions.
    \item We develop alternating optimization of planner and aggregator LoRA adapters, aligning plans with executor capabilities and candidate generation with selection.
    \item Experiments across five benchmarks and three backbones demonstrate improved accuracy--token trade-offs compared with training-free and training-based baselines.
\end{itemize}

%% file: sections/motivation_figures.tex
\begin{figure}[t]
\centering
\includegraphics[width=0.95\linewidth]{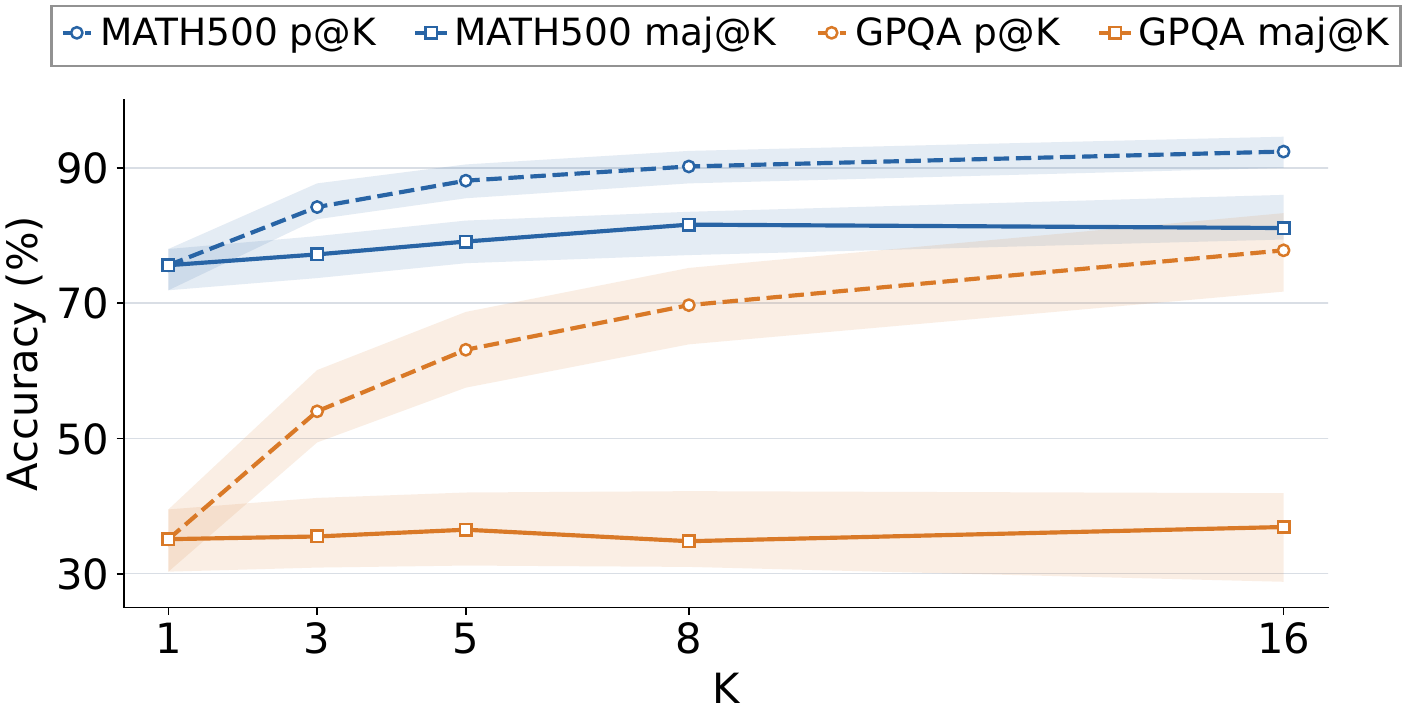}
\caption{Majority-vote accuracy and candidate coverage under repeated solution sampling with Qwen2.5-7B-Instruct on MATH500 and GPQA Diamond. Shaded regions show confidence intervals.}
\label{fig:no_plan_scaling}

\vspace{8pt}
\includegraphics[width=0.95\linewidth]{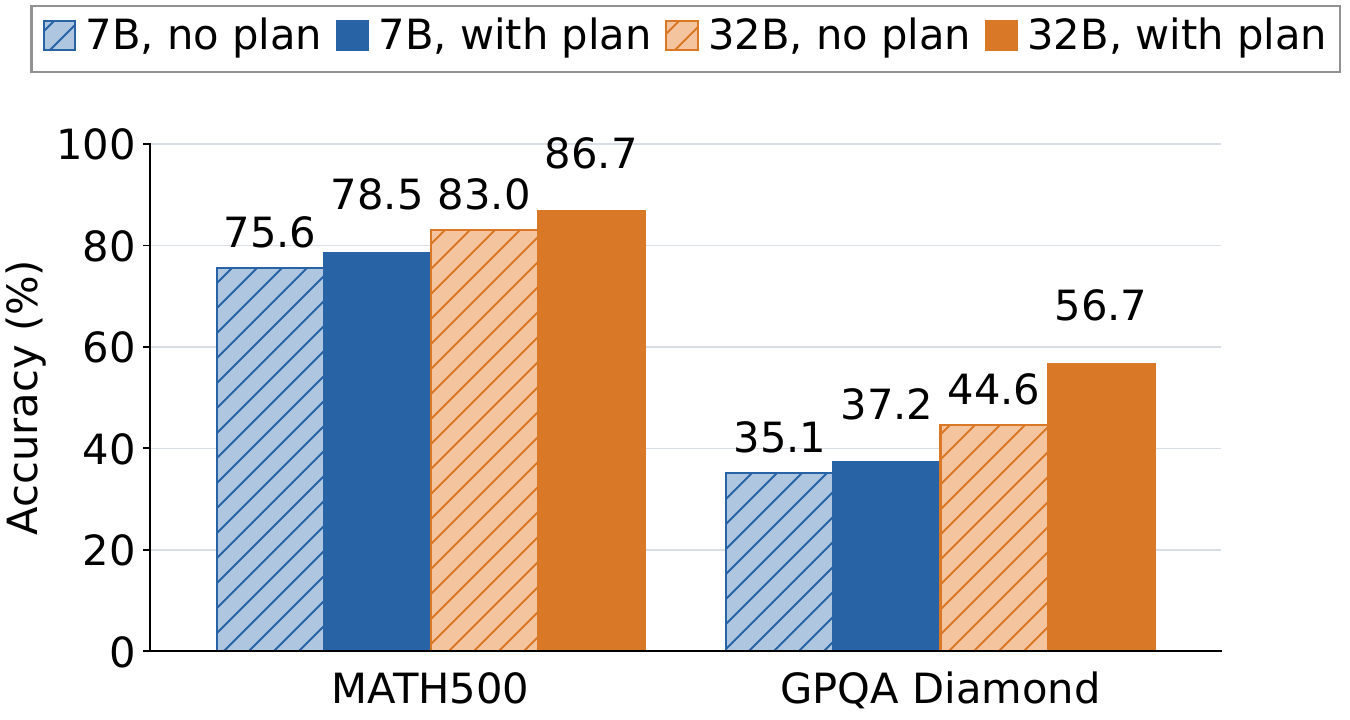}
\caption{Accuracy of Qwen2.5-7B-Instruct and Qwen2.5-32B-Instruct with and without plans. Shared GPT-5.6 Sol plans are limited to 200 tokens and checked for answer leakage; both executors are verified to generate intermediate reasoning before answering.}
\label{fig:executor_plan_accuracy}
\end{figure}

%% file: sections/framework_figure.tex
\begin{figure*}[t]
\centering
\includegraphics[width=0.9\textwidth]{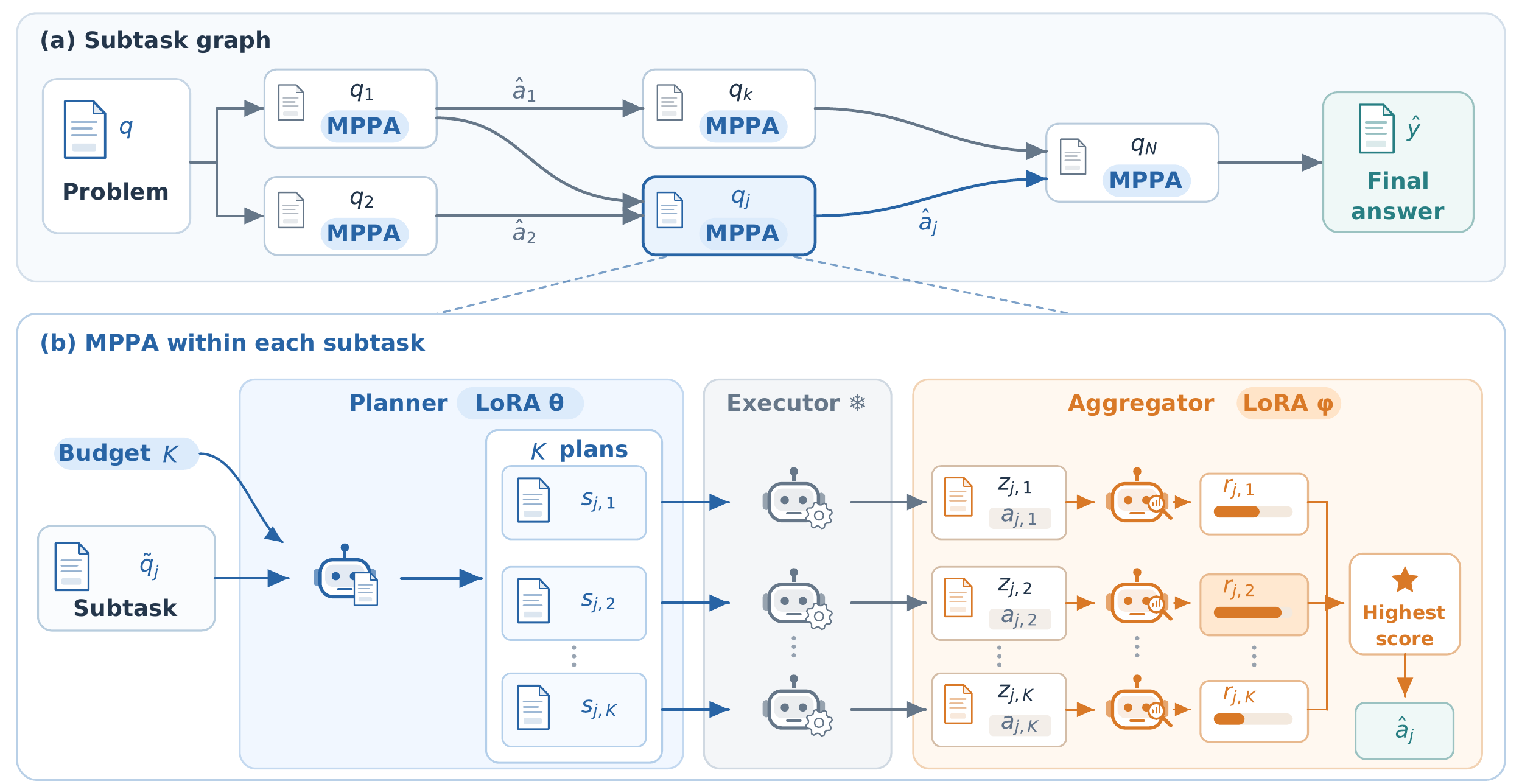}
\caption{Overview of \methodname{}. (a) Selected answers propagate along task-provided subtask dependencies. (b) Within each subtask, the planner jointly generates plans, the frozen executor solves once per plan, and the aggregator selects the highest-scoring candidate.}
\label{fig:mppa_framework}
\end{figure*}

%% file: sections/motivation.tex
We examine how additional inference-time computation translates into correct final answers for a frozen executor. Specifically, we study two aspects of test-time scaling: how candidate solutions are generated and how the final answer is selected. These observations inform the design of the planner and aggregator.

\noindent \textbf{Majority voting leaves substantial selection headroom.} We first evaluate repeated solution sampling without plans. For each problem, we generate multiple candidate solutions and compare majority-vote accuracy (maj@$K$) with candidate coverage (pass@$K$), the fraction of problems with at least one correct solution among $K$ samples. As shown in \Cref{fig:no_plan_scaling}, increasing $K$ raises candidate coverage on both datasets, but the gains in majority-vote accuracy are substantially smaller. On MATH500, voting accuracy improves gradually, while on GPQA Diamond it remains nearly unchanged despite the growing availability of correct candidates. Additional samples often contain a correct solution without changing the majority answer. This motivates learning to select answers using reasoning evidence beyond answer frequency.

\noindent\textbf{Plan utility depends on executor capability.} We evaluate Qwen2.5-7B-Instruct and Qwen2.5-32B-Instruct on MATH500 and GPQA Diamond using identical plans generated by GPT-5.6 Sol~\citep{openai2026gpt56}. We check the plans for answer leakage and verify that both executors generate intermediate reasoning before giving a final answer. We limit each plan to at most 200 tokens. As shown in \Cref{fig:executor_plan_accuracy}, the same plans provide greater benefits to the stronger executor, particularly on GPQA Diamond.
Case analyses reveal errors in intermediate calculations, problem interpretation, and adherence to conditions stated in the plan. Effective plans should therefore account for the executor's capabilities, offering approaches that it can reliably carry out.

%% file: sections/method.tex
\subsection{Overview}
\label{sec:planning_execution}

A task $q$ is supplied with $N$ task-provided subtasks $q_1,\ldots,q_N$ and their dependencies. When no subtasks are specified, we set $N=1$ and treat the original task as the sole subtask. As illustrated in \Cref{fig:mppa_framework}, \methodname{} comprises a planner $P_\theta$, a frozen executor $E$, and an aggregator $A_\phi$. All three components share a frozen language-model backbone, with separate LoRA adapters~\citep{hu2022lora} for planning and aggregation. The executor uses the backbone with both adapters disabled.

For each subtask $q_j$, we incorporate the selected answers from its prerequisites to form the current input $\tilde q_j$. We use the same plan budget $K$ for all subtasks of a task. The planner generates an ordered collection of plans in a single call:
\begin{equation}
    S_j=(s_{j,1},\ldots,s_{j,K})
    \sim P_\theta(\cdot\mid\tilde q_j,K).
    \label{eq:plan_set}
\end{equation}
Each plan outlines an approach to solving the full subtask, leaving detailed derivations and the answer to the executor. Joint generation allows later plans to condition on previously proposed approaches, encouraging complementary routes. Conditioning on $K$ supports different inference budgets with the same planner. The frozen executor then solves the subtask once under each plan:
\begin{equation}
    z_{j,i}\sim E(\cdot\mid\tilde q_j,s_{j,i}),
    \quad i=1,\ldots,K.
    \label{eq:plan_execution}
\end{equation}
These executions can run in parallel, producing $K$ complete candidate solutions.

The aggregator scores each candidate given $\tilde q_j$ and selects the answer from the highest-scoring candidate. Selected answers are passed to dependent subtasks, and the resulting subtask outputs are combined as required to obtain the final task answer. The complete task-level inference procedure is summarized in \Cref{alg:mppa_inference} in \Cref{app:inference}. We train the planner and aggregator on individual subtasks, with rewards and supervision defined against each subtask's reference answer. When gold subtask answers are unavailable, we sample multiple complete teacher traces for the task and retain only traces with correct final task answers. We then obtain a reference answer for each subtask by majority voting over its answers in the retained traces.

\subsection{Learning the Multi-Path Planner}
\label{sec:planner_learning}

\paragraph{Supervised initialization.}
\label{SFT_data}
We initialize the planner $P_\theta$ through supervised fine-tuning on teacher-generated plan sets (\Cref{app:supervision_construction}). Each training example pairs a subtask input $\tilde q_j$ and a budget $K$ with an ordered collection of $K$ plans. Training across different budgets enables the same planner to generate a variable number of paths. We limit each plan to at most 200 tokens to keep the guidance concise and limit planning overhead, and check the plans for answer leakage (\Cref{app:prompts}). We optimize the likelihood of the complete plan collection:
\begin{equation}
    \mathcal L_{\mathrm{SFT}}
    =-\mathbb E_{(\tilde q_j,K,S_j)}
    \log P_\theta(S_j\mid\tilde q_j,K).
    \label{eq:planner_sft}
\end{equation}

\paragraph{On-policy optimization.}
We further optimize the planner using the outcomes of frozen execution and learned aggregation. Let $c_{j,i}\in\{0,1\}$ indicate whether candidate $z_{j,i}$ is correct, and let $i^\star=\arg\max_i r_{j,i}$, where $r_{j,i}=A_\phi(\tilde q_j,z_{j,i})$. We assign the plan set a reward
\begin{equation}
    R(S_j)
    =\lambda c_{j,i^\star}
    +(1-\lambda)\max_{1\le i\le K}c_{j,i},
    \label{eq:planner_reward}
\end{equation}
where $\lambda\in[0,1]$. The first term rewards correctness after aggregation, aligning planning with the deployed selection procedure. The second rewards candidate coverage, providing a correctness signal independent of the aggregator's current ranking. Together, these terms evaluate plans through the solutions that the executor can produce under their guidance.

For each training input, we sample $G$ plan sets using the same budget $K$, execute their plans, and compute their rewards. We normalize rewards within each input--budget group by subtracting the group mean, without standard-deviation scaling, and update the planner LoRA with GRPO~\citep{shao2024deepseekmath}. Each sampled plan collection is treated as a single action, so optimization operates on the collection generated in one call. The executor and aggregator remain fixed during each planner update (\Cref{app:optimization_details}).

\begin{table*}[t]
\centering
\small
\caption{Accuracy (\%) and output tokens relative to Base. GPQA and MMLU-Pro denote GPQA Diamond and MMLU-Pro-sci500. Column-wise best accuracies within each model are bold.}
\label{tab:main_results}
\renewcommand{\arraystretch}{0.95}
\resizebox{0.7\linewidth}{!}{%
\begin{tabular}{@{}lcccccrr@{}}
\toprule
\multirow[c]{2}{*}[-0.6ex]{Method} & \multicolumn{3}{c}{Mathematics} & \multicolumn{2}{c}{Science} & \multirow[c]{2}{*}[-0.6ex]{Avg} & \multirow[c]{2}{*}[-0.6ex]{Token} \\
\cmidrule(lr){2-4}\cmidrule(lr){5-6}
 & MATH500 & AIME24 & AIME25 & GPQA & MMLU-Pro & & \\
\midrule
\multicolumn{8}{c}{Qwen2.5-7B-Instruct} \\
\midrule
Base & 75.6 & 12.6 & 6.7 & 35.1 & 64.6 & 38.9 & $1.00\times$ \\
Base maj@8 & 81.6 & 14.4 & 13.3 & 34.8 & 69.6 & 42.8 & $8.00\times$ \\
PS+ maj@8 & 80.6 & 15.6 & 10.0 & 37.9 & 69.6 & 42.7 & $8.39\times$ \\
L2M maj@8 & 80.4 & 18.9 & 11.1 & 33.8 & 71.8 & 43.2 & $32.42\times$ \\
ESC ($w=2$) & 82.0 & 14.4 & 18.9 & 37.4 & 70.6 & 44.7 & $4.94\times$ \\
USC & 80.2 & 15.6 & 15.6 & 34.3 & 66.4 & 42.4 & $8.09\times$ \\
Base agg@8 & 87.9 & 21.1 & 20.0 & 69.4 & 86.7 & 57.0 & $8.00\times$ \\
SFT agg@8 & 83.6 & 17.8 & 16.7 & \textbf{80.3} & \textbf{87.6} & 57.2 & $15.50\times$ \\
GRPO agg@8 & 88.2 & 21.1 & 18.9 & 70.7 & 85.6 & 56.9 & $8.94\times$ \\
\rowcolor[gray]{0.94}
MPPA @1 & 83.6 & 16.7 & 14.4 & 45.1 & 71.9 & 46.3 & $1.38\times$ \\
\rowcolor[gray]{0.94}
MPPA @3 & \textbf{89.2} & \textbf{22.2} & \textbf{24.4} & 72.2 & 84.0 & \textbf{58.4} & $4.23\times$ \\
\midrule
\multicolumn{8}{c}{Gemma-3-12B-it} \\
\midrule
Base & 85.6 & 21.9 & 18.2 & 39.7 & 66.8 & 46.4 & $1.00\times$ \\
Base maj@8 & 89.2 & 27.8 & 21.1 & 44.9 & 72.6 & 51.1 & $8.00\times$ \\
PS+ maj@8 & 88.6 & 28.9 & 21.1 & 42.4 & 76.0 & 51.4 & $8.08\times$ \\
L2M maj@8 & 89.4 & 30.0 & 18.9 & 45.5 & 74.6 & 51.7 & $19.71\times$ \\
ESC ($w=2$) & 88.0 & 31.1 & 24.4 & 43.9 & 71.4 & 51.8 & $4.44\times$ \\
USC & 87.0 & 25.6 & 22.2 & 40.9 & 66.0 & 48.3 & $8.32\times$ \\
Base agg@8 & 92.8 & 34.4 & \textbf{31.1} & 73.7 & 87.2 & 63.9 & $8.00\times$ \\
SFT agg@8 & 85.0 & 23.3 & 24.4 & \textbf{84.8} & \textbf{89.0} & 61.3 & $12.18\times$ \\
GRPO agg@8 & \textbf{93.4} & 37.8 & \textbf{31.1} & 71.2 & 87.2 & 64.1 & $7.90\times$ \\
\rowcolor[gray]{0.94}
MPPA @1 & 88.3 & 40.0 & 18.9 & 49.3 & 75.8 & 54.5 & $1.28\times$ \\
\rowcolor[gray]{0.94}
MPPA @3 & 93.2 & \textbf{52.2} & \textbf{31.1} & 70.7 & 83.8 & \textbf{66.2} & $3.90\times$ \\
\midrule
\multicolumn{8}{c}{Phi-4 (14B)} \\
\midrule
Base & 80.6 & 16.7 & 18.1 & 54.2 & 79.4 & 49.8 & $1.00\times$ \\
Base maj@8 & 84.4 & 21.1 & 23.3 & 59.1 & 83.6 & 54.3 & $8.00\times$ \\
PS+ maj@8 & 84.2 & 22.2 & 20.0 & 57.6 & 83.2 & 53.4 & $8.15\times$ \\
L2M maj@8 & 85.8 & 22.2 & 21.1 & 61.1 & 84.4 & 54.9 & $30.98\times$ \\
ESC ($w=2$) & 84.2 & 22.2 & 18.9 & 60.1 & 85.0 & 54.1 & $4.62\times$ \\
USC & 83.6 & 18.9 & 22.2 & 59.6 & 81.6 & 53.2 & $8.24\times$ \\
Base agg@8 & 91.0 & 31.1 & \textbf{31.1} & 81.8 & 91.8 & 65.4 & $8.00\times$ \\
SFT agg@8 & 89.8 & 37.8 & 30.0 & \textbf{86.4} & \textbf{93.4} & \textbf{67.5} & $11.12\times$ \\
GRPO agg@8 & 91.4 & 33.3 & 24.4 & 85.9 & 91.6 & 65.3 & $8.01\times$ \\
\rowcolor[gray]{0.94}
MPPA @1 & 87.2 & 26.7 & 24.4 & 65.0 & 84.1 & 57.5 & $1.28\times$ \\
\rowcolor[gray]{0.94}
MPPA @3 & \textbf{92.4} & \textbf{40.0} & \textbf{31.1} & 82.3 & 90.4 & 67.2 & $3.67\times$ \\
\bottomrule
\end{tabular}
}
\end{table*}

\subsection{Learning the Candidate Aggregator}
\label{sec:aggregator_learning}

\paragraph{Pointwise scoring.}
The aggregator $A_\phi$ evaluates each candidate solution $z_{j,i}$ independently, conditioned on the subtask input $\tilde q_j$. At the scoring prompt's answer position, we compute the difference between the logits of the output tokens \texttt{Yes} and \texttt{No}:
\begin{equation}
    r_{j,i}=\ell_{j,i}^{\mathrm{Yes}}-\ell_{j,i}^{\mathrm{No}}.
    \label{eq:aggregator_score}
\end{equation}
The corresponding correctness probability is $\sigma(r_{j,i})$, where $\sigma$ is the sigmoid function. The same logit difference is used for ranking at inference time. Independent scoring supports variable plan budgets while keeping each input limited to one candidate solution.

\paragraph{Two-stage training.}
Listwise supervision is available only for groups containing both correct and incorrect candidates, limiting the amount of training data and weakening the early training signal. We therefore first initialize the aggregator using pointwise binary cross-entropy on the correctness labels $c_{j,i}$ across all candidates, providing a stronger starting point to stabilize subsequent ranking optimization:
\begin{equation}
    \mathcal L_{\mathrm{pt}}
    =\mathbb E_{j,i}\!\left[
        \operatorname{BCE}\bigl(\sigma(r_{j,i}),c_{j,i}\bigr)
    \right].
    \label{eq:aggregator_pointwise}
\end{equation}
Starting from this adapter, we continue training with a combined pointwise and listwise objective. Each list comprises the $K$ candidates generated for the same subtask input. Let $\mathcal G_{\mathrm{mix}}$ denote the indices of groups containing both correct and incorrect candidates. The listwise loss rewards the total softmax probability assigned to correct candidates:
\begin{equation}
    \mathcal L_{\mathrm{list}}
    =-\mathbb E_{j\in\mathcal G_{\mathrm{mix}}}\!
    \log\frac{\sum_{i:c_{j,i}=1}\exp(r_{j,i})}
              {\sum_{i=1}^{K}\exp(r_{j,i})}.
    \label{eq:aggregator_listwise}
\end{equation}
The second-stage objective is
\begin{equation}
    \mathcal L_{\mathrm{agg}}
    =\mathcal L_{\mathrm{pt}}+\eta\mathcal L_{\mathrm{list}},
    \label{eq:aggregator_joint}
\end{equation}
where the pointwise term is averaged over all candidates and the listwise term is averaged over groups containing both correct and incorrect candidates, $\mathcal G_{\mathrm{mix}}$. We initially set $\eta=1$ and use a lower learning rate and fewer update steps in the second stage. Both stages update only the aggregator LoRA; the shared backbone remains frozen.

\input{sections/ablation_tables.tex}

\subsection{Alternating Optimization}
\label{sec:alternating_optimization}

\paragraph{Initialization.}
We initialize the planner with SFT (\Cref{sec:planner_learning}), then train the aggregator with pointwise supervision on candidates generated by this planner and the frozen executor (\Cref{sec:aggregator_learning}).

\paragraph{Shared data collection.}
Each round samples subtask inputs and one budget $K$ per input, then generates $G$ plan sets at that budget. We execute each plan once and collect candidates, correctness labels, and aggregator scores for both updates.

\paragraph{Alternating updates.}
We update the planner with the aggregator fixed, then update the aggregator with the planner fixed using the joint loss and replay data from recent rounds (\Cref{sec:planner_learning,sec:aggregator_learning}). Since planner rewards depend on the aggregator and aggregator training data come from the planner, we use short alternating updates to limit abrupt changes in their interaction. Both updates reuse the collected candidates without additional executor calls. We retain the pair with the best held-out validation performance across plan budgets. The complete procedure appears in \Cref{alg:mppa_training} (\Cref{app:training}).

%% file: sections/ablation_tables.tex
\begin{table*}[t]
\centering
\small
\caption{Generation and selection ablations on Qwen2.5-7B-Instruct with the fixed number of candidates $K=3$. Single plan reuses one plan across $K$ executions; independent plans use $K$ separate planning calls.}
\label{tab:generation_selection_ablations}
\resizebox{0.75\linewidth}{!}{%
\begin{tabular}{@{}llcccccr@{}}
\toprule
\multirow[c]{2}{*}[-0.6ex]{Generation} & \multirow[c]{2}{*}[-0.6ex]{Selection} & \multicolumn{3}{c}{Mathematics} & \multicolumn{2}{c}{Science} & \multirow[c]{2}{*}[-0.6ex]{Avg} \\
\cmidrule(lr){3-5}\cmidrule(lr){6-7}
 & & MATH500 & AIME24 & AIME25 & GPQA & MMLU-Pro & \\
\midrule
No plan & Majority vote & 77.2 & 13.3 & 7.8 & 35.5 & 67.3 & 40.2 \\
No plan & Aggregator & 83.6 & 17.8 & 13.3 & 54.0 & 78.3 & 49.4 \\
Single plan & Aggregator & 88.0 & 18.9 & 16.7 & 67.2 & 80.1 & 54.2 \\
Independent $K$ plans & Aggregator & 87.2 & 20.0 & 13.3 & 69.0 & 81.4 & 54.2 \\
Joint $K$ plans & Majority vote & 84.8 & 15.6 & 17.8 & 46.5 & 74.2 & 47.8 \\
\rowcolor[gray]{0.94}
Joint $K$ plans & Aggregator & \textbf{89.2} & \textbf{22.2} & \textbf{24.4} & \textbf{72.2} & \textbf{84.0} & \textbf{58.4} \\
\bottomrule
\end{tabular}%
}
\end{table*}

\begin{table*}[t]
\centering
\small
\caption{Training ablations on Qwen2.5-7B-Instruct. Stage 1 uses pointwise supervision; Stage 2 adds listwise training. Alternating optimization updates both adapters for $T=3$ rounds.}
\label{tab:training_ablations}
\resizebox{0.75\linewidth}{!}{%
\begin{tabular}{@{}llcccccr@{}}
\toprule
\multirow[c]{2}{*}[-0.6ex]{Planner} & \multirow[c]{2}{*}[-0.6ex]{Aggregator} & \multicolumn{3}{c}{Mathematics} & \multicolumn{2}{c}{Science} & \multirow[c]{2}{*}[-0.6ex]{Avg} \\
\cmidrule(lr){3-5}\cmidrule(lr){6-7}
 & & MATH500 & AIME24 & AIME25 & GPQA & MMLU-Pro & \\
\midrule
SFT & Stage 1 & 80.5 & 14.4 & 10.0 & 39.6 & 70.0 & 42.9 \\
SFT $\rightarrow$ RL & Stage 1 & 83.6 & 16.7 & 12.2 & 43.4 & 73.2 & 45.8 \\
SFT $\rightarrow$ RL & Stage 1 $\rightarrow$ Stage 2 & 86.6 & 18.9 & 15.6 & 46.0 & 76.2 & 48.6 \\
\rowcolor[gray]{0.94}
\multicolumn{2}{@{}l}{Alternating optimization ($T=3$)} & \textbf{89.2} & \textbf{22.2} & \textbf{24.4} & \textbf{72.2} & \textbf{84.0} & \textbf{58.4} \\
\bottomrule
\end{tabular}%
}
\end{table*}

%% file: sections/experiments.tex
\begin{figure*}[t]
\centering
\captionsetup{skip=2pt}
\begin{minipage}[t]{0.325\textwidth}
\centering
\includegraphics[width=\linewidth,trim=0 6bp 0 0,clip]{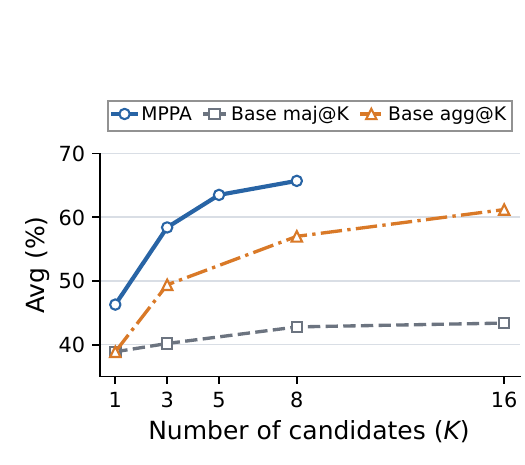}
\caption{Average accuracy on five benchmarks by number of candidates for Qwen2.5-7B-Instruct.}
\label{fig:macro_accuracy_scaling}
\end{minipage}
\hfill
\begin{minipage}[t]{0.325\textwidth}
\centering
\includegraphics[width=\linewidth,trim=0 6bp 0 0,clip]{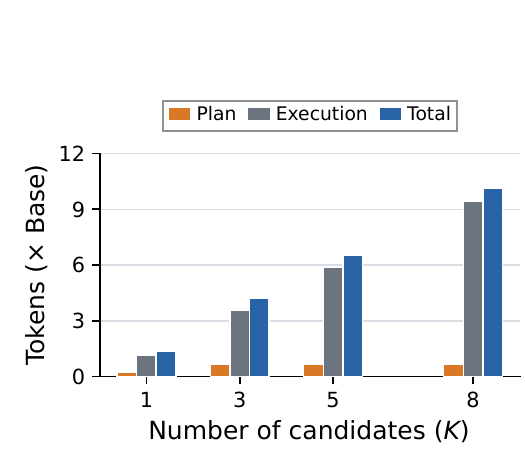}
\caption{MPPA output tokens from planning, execution, and both combined, averaged across five benchmarks, using Qwen2.5-7B-Instruct.}
\label{fig:token_usage_scaling}
\end{minipage}
\hfill
\begin{minipage}[t]{0.325\textwidth}
\centering
\includegraphics[width=\linewidth,trim=0 6bp 0 0,clip]{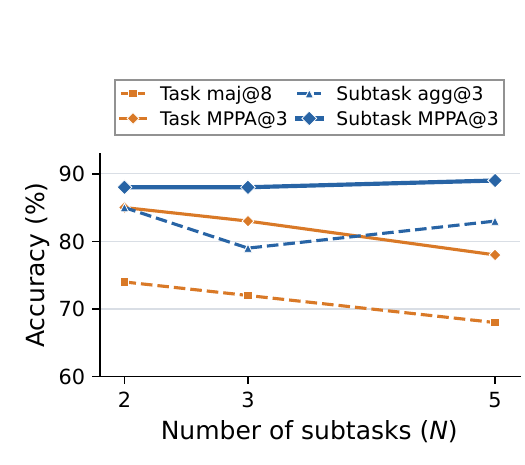}
\caption{Final-task accuracy by number of provided subtasks on CompMATH, using Qwen2.5-7B-Instruct.}
\label{fig:subtask_accuracy_by_n}
\end{minipage}
\end{figure*}

\begin{table*}[t]
\centering
\begin{minipage}[t]{0.54\textwidth}
\centering
\caption{Final-task accuracy (\%) and output tokens (relative to task Base pass@1) on CompMATH, using Qwen2.5-7B-Instruct.}
\label{tab:subtask_results}
\footnotesize
\setlength{\tabcolsep}{3pt}
\renewcommand{\arraystretch}{1}
\resizebox{\linewidth}{!}{%
\begin{tabular}{@{}lccccc@{}}
\toprule
Method & $N=2$ & $N=3$ & $N=5$ & Avg & Tokens ($\times$) \\
\midrule
Task Base pass@1 & 64.2 & 58.2 & 55.2 & 59.2 & 1.00 \\
Task Base maj@8 & 74.0 & 72.0 & 68.0 & 71.3 & 7.95 \\
Task Base agg@8 & 74.0 & 79.0 & 66.0 & 73.0 & 7.95 \\
\rowcolor[gray]{0.94}
Task MPPA@3 & 85.0 & 83.0 & 78.0 & 82.0 & 3.64 \\
\midrule
Subtask Base maj@3 & 84.0 & 73.0 & 79.0 & 78.7 & 5.34 \\
Subtask Base agg@3 & 85.0 & 79.0 & 83.0 & 82.3 & 5.34 \\
\rowcolor[gray]{0.94}
Subtask MPPA@3 & \textbf{88.0} & \textbf{88.0} & \textbf{89.0} & \textbf{88.3} & 6.98 \\
\bottomrule
\end{tabular}%
}
\end{minipage}
\hfill
\begin{minipage}[t]{0.4\textwidth}
\centering
\caption{Subtask accuracy (\%) on CompMATH, using Qwen2.5-7B-Instruct. Cond. Subtask Acc. considers only subtasks whose prerequisite answers are all correct.}
\label{tab:subtask_accuracy}
\footnotesize
\setlength{\tabcolsep}{3pt}
\renewcommand{\arraystretch}{0.8}
\resizebox{\linewidth}{!}{%
\begin{tabular}{@{}c|lrr@{}}
\toprule
$N$ & Method & \begin{tabular}[c]{@{}r@{}}Subtask\\Acc.\end{tabular} & \begin{tabular}[c]{@{}r@{}}Cond. Sub-\\task Acc.\end{tabular} \\
\midrule
\multirow[c]{3}{*}{$2$} & Subtask Base maj@3 & 87.5 & 91.6 \\
 & Subtask Base agg@3 & 87.5 & 91.6 \\
\rowcolor[gray]{0.94}
\cellcolor{white} & Subtask MPPA@3 & \textbf{91.0} & \textbf{93.8} \\
\midrule
\multirow[c]{3}{*}{$3$} & Subtask Base maj@3 & 85.3 & 90.4 \\
 & Subtask Base agg@3 & 89.7 & 92.8 \\
\rowcolor[gray]{0.94}
\cellcolor{white} & Subtask MPPA@3 & \textbf{93.0} & \textbf{95.9} \\
\midrule
\multirow[c]{3}{*}{$5$} & Subtask Base maj@3 & 89.8 & 95.1 \\
 & Subtask Base agg@3 & 92.6 & 96.5 \\
\rowcolor[gray]{0.94}
\cellcolor{white} & Subtask MPPA@3 & \textbf{94.2} & \textbf{97.1} \\
\bottomrule
\end{tabular}%
}
\end{minipage}
\end{table*}

\subsection{Experimental Setup}
\label{sec:experimental_setup}

\paragraph{Datasets.}
We evaluate on the MATH500 test set~\citep{hendrycks2021measuring,lightman2023let}, AIME 2024 and 2025~\citep{AIME2024}, GPQA Diamond~\citep{gpqa}, and MMLU-Pro-sci500~\citep{wang2024mmlupro}. The latter contains 500 questions randomly sampled from the physics, chemistry, and biology portions of MMLU-Pro in proportion to their respective sizes. 
We construct a shared training pool of 18,319 mathematics and science problems.
Detailed composition, filtering, and deduplication procedures are provided in \Cref{app:datasets}.
These benchmarks do not provide subtasks, so we treat each problem as a single subtask ($N=1$) and apply \methodname{} to the complete problem. We separately evaluate datasets with task-provided subtasks later in this section.
We report macro-average accuracy (Avg) across the five datasets.
For all reported results, we repeat each evaluation three times with independent random seeds and report the mean.

\paragraph{Models.}
We use Qwen2.5-7B-Instruct~\citep{yang2024qwen2}, Gemma-3-12B-it~\citep{gemmateam2025gemma3}, and Phi-4 (14B)~\citep{abdin2024phi4} as backbone models (See \Cref{app:training_configurations}).

\paragraph{Baselines.}
We compare against the original model checkpoint (Base), supervised fine-tuning on complete solutions (SFT), and reinforcement learning with GRPO (GRPO)~\citep{shao2024deepseekmath}. We also evaluate Plan-and-Solve Prompting (PS+)~\citep{wang2023planandsolve} and Least-to-Most Prompting (L2M)~\citep{zhou2023leasttomost}, which introduce planning and decomposition through prompting; Early-Stopping Self-Consistency (ESC)~\citep{li2024earlystopping}, which stops sampling when answers agree within a window; and Universal Self-Consistency (USC)~\citep{chen2023universal}, which uses the model to select the most consistent response from multiple candidates. Base reports single-sample accuracy, while Base maj@8, PS+ maj@8, and L2M maj@8 use majority voting over eight sampled solutions. Base agg@8, SFT agg@8, and GRPO agg@8 use the learned aggregator to select from eight solutions generated by the corresponding model. USC also selects from eight candidates. ESC uses a window size of $w=2$ and a maximum of eight samples (See \Cref{app:baseline_implementations}).

\subsection{Main Results}
\label{sec:main_results}

\Cref{tab:main_results} suggests that effective test-time scaling depends on how candidates are generated and selected. The strong performance of Base agg@8 and SFT agg@8 highlights the value of learned selection. MPPA further improves the accuracy--cost trade-off by learning plans tailored to the frozen executor: MPPA@3 achieves the highest average accuracy on Qwen2.5-7B and Gemma-3-12B, and closely matches SFT agg@8 on Phi-4 with roughly one-third of its output tokens. Moreover, MPPA@1 surpasses Base maj@8 in average accuracy across all three backbones, indicating that learned planning can make a single execution more effective than repeated unguided sampling (\Cref{tab:main_results_intervals}).

\subsection{Analysis}
\label{sec:analysis}
\label{sec:ablation_studies}
\label{sec:generation_selection_ablations}

\paragraph{Generation and selection.}
\Cref{tab:generation_selection_ablations} shows that candidate generation and answer selection provide complementary gains with Qwen2.5-7B-Instruct at $K=3$. The learned aggregator improves accuracy with both unguided sampling and joint planning, showing that majority voting leaves useful candidates unselected. With the aggregator fixed, joint planning outperforms both reusing a single plan and sampling plans independently. This suggests that generating plans together provides more effective candidate sets for aggregation. The strongest performance comes from combining joint planning with learned selection (\Cref{tab:candidate_coverage_selection}).

\paragraph{Training.}
\Cref{tab:training_ablations} shows progressive gains from planner reinforcement learning, the aggregator's second training stage, and alternating optimization. Planner RL optimizes plans for downstream execution and selection (\Cref{fig:grpo_training_curve}), while listwise training aligns aggregator scores with competition among candidates. Alternating updates achieve the strongest performance, supporting the value of adapting the planner and aggregator together as the candidate distribution changes.

\paragraph{Accuracy and cost scaling.}
\label{sec:budget_scaling}
\Cref{fig:macro_accuracy_scaling} shows that MPPA makes more effective use of additional candidates than direct sampling. MPPA with five plans exceeds the average accuracy of Base agg@16, indicating that learned planning improves the value of each execution even when both methods use learned selection. \Cref{fig:token_usage_scaling} shows that execution accounts for most output tokens, while planning contributes a small fraction. Together, these results indicate that the planning overhead is offset by requiring fewer executions to reach comparable accuracy (\Cref{fig:dataset_accuracy_cost}).

\subsection{Tasks with Provided Subtasks}
\label{sec:subtask_results}

We construct CompMATH, a composite mathematics benchmark derived from MATH500 with task-provided subtasks and dependencies. For each $N\in\{2,3,5\}$, we sample approximately 100 distinct seed problems and prescribe dependency graphs (a chain for $N=2$, fan-in or triangular graphs for $N=3$, and branching DAGs such as diamonds for $N=5$). GPT-5.6 Sol generates problems and decompositions with numerical reference answers; we retain examples passing structural checks and answer-aware review by GPT-5.6 Terra~\citep{openai2026gpt56}. Task-level methods receive the complete problem, while subtask-level methods also receive its decomposition. Construction details appear in \Cref{app:composite_tasks}.

\Cref{fig:subtask_accuracy_by_n} shows that Subtask MPPA@3 maintains high final-task accuracy as the number of subtasks grows, while Task MPPA@3 declines. It also outperforms Subtask Base agg@3 under the same decomposition and execution budget (\Cref{tab:subtask_results}), demonstrating the benefit of learned planning. Its higher accuracy on subtasks with correct prerequisites (\Cref{tab:subtask_accuracy}) helps produce more reliable answers downstream (\Cref{app:example_outputs}).

%% file: sections/related_work.tex
\paragraph{Test-time scaling.}
Test-time scaling improves reasoning by allocating additional computation during inference through serial or parallel expansion. Serial scaling extends computation within a single reasoning trajectory, as exemplified by OpenAI o1~\citep{openai2024reasoning} and DeepSeek-R1~\citep{guo2025deepseek}. s1 further enables controllable serial scaling through budget forcing~\citep{muennighoff2025s1simpletesttimescaling}. Parallel scaling generates multiple candidate solutions and selects or aggregates their outputs through self-consistency or verifier-based best-of-$N$ selection~\citep{wang2023selfconsistency,cobbe2021trainingverifiers}. Self-Certainty offers a more recent approach to best-of-$N$ selection using the model's output distributions without an external reward model~\citep{kang2025selfcertainty}. Large Language Monkeys shows that repeated sampling substantially increases candidate coverage, while answer selection can limit the resulting accuracy gains~\citep{brown2024monkeys}. Studies of inference scaling laws and compute-optimal inference further examine how strategies, model capabilities, and problem difficulty shape the accuracy--cost trade-off~\citep{wu2024inferencescaling,snell2024scaling}. Related work also explores adaptive orchestration and uncertainty-aware agent learning~\citep{su2026difficulty,lan2026trust,lin2026uncertainty,zhang2026uopd}. \methodname{} contributes to this direction by learning coordinated multi-path planning and candidate selection around a frozen executor, improving accuracy at a given budget.

\paragraph{Planning for reasoning.}
Research on planning for reasoning follows two complementary directions: structuring execution and diversifying solution paths~\citep{hao2023reasoning,xiong2024deliberate}. The first separates high-level guidance from detailed reasoning. Plan-and-Solve generates a plan before solving~\citep{wang2023planandsolve}, while Least-to-Most decomposes problems into sequential subproblems~\citep{zhou2023leasttomost}. MLR interleaves planning and execution, adapting plans to execution progress~\citep{xiong2026mlr}. TraceLift learns planning strategies from reasoning quality and utility for a frozen executor~\citep{han2026tracelift}. Related work explores structured reasoning across symbolic and specialized reasoning settings~\citep{yang2024can,yang2024harnessing,xiong2024large,xiong2025symplanner,he2024give,he2025selfgiveassociativethinkinglimited,yu2025causaleval}. The second direction uses planning to broaden exploration. PlanSearch explores diverse natural-language plans for code generation~\citep{wang2024plansearch}, while PTTS jointly generates branch outlines and optimizes candidate coverage through pass@$K$ rewards~\citep{wu2026ptts}. Multi-path and multi-agent approaches coordinate multiple reasoning processes across diverse applications~\citep{he2024enhancing,luo2025llm,luo2026resisting,su2026swingguard}. Recent work also explores self-evolving and adaptive agentic reasoning~\citep{wang2026reflex,wang2026atlasva,hu2026seal,liu2026palm,shen2026fine}. \methodname{} combines coordinated planning with learned candidate selection. Alternating planner and aggregator optimization aligns exploration with final selection using rewards for selected-answer correctness and candidate coverage.

\paragraph{Verification and answer selection.}
Outcome-supervised verifiers rank complete solutions by predicted correctness to improve best-of-$N$ selection~\citep{cobbe2021trainingverifiers}. Process-supervised verifiers assess intermediate reasoning steps using human annotations~\citep{lightman2023let} or automatically constructed labels~\citep{wang2023math}. Qwen2.5-Math-PRM further improves supervision through consensus filtering of Monte Carlo estimates and LLM judgments~\citep{zhang2025prmlessons}. Generative verification offers another approach: GenRM models correctness through next-token prediction and supports explicit verification reasoning~\citep{zhang2025genrm}. Methods exploiting information across candidates include Universal Self-Consistency, which selects the most consistent response using an LLM~\citep{chen2023universal}, and Multi-Sequence Verifiers, which predict correctness conditioned on the full candidate set~\citep{kim2026msv}. Recent agent research also studies verification and correction in memory-augmented agents~\citep{RollbackRepair,EpistemicAdmission,PACMI,li2026remember}. Related systems also apply structured search and selection to scientific discovery, information retrieval, and agent robustness~\citep{li2026learning,hu2026beyond,ma-etal-2026-intragent,rao2026fragfusebypassingaccesscontrol}. \methodname{} alternates planner and aggregator optimization, combining pointwise and listwise supervision for independent candidate scoring with planning rewards based on execution and selection outcomes.

%% file: sections/conclusion.tex
We introduced \methodname{}, which learns coordinated multi-path planning and answer selection around a frozen executor. Alternating optimization aligns plans with the executor and adapts the aggregator to the resulting candidates. Experiments across math and science benchmarks show improved accuracy--token trade-offs over training-free and training-based baselines. These results highlight the value of jointly creating useful solution paths and selecting correct answers. Future work will explore automatic task decomposition and adaptive plan-budget allocation across subtasks.

%% file: sections/limitations.tex
MPPA improves planning and answer selection while keeping the executor frozen. Further gains may be possible by also optimizing the execution process, such as adapting the executor to better follow plans or allocating computation within each execution. In addition, our method and evaluation focus on text-based reasoning. Extending MPPA to multimodal tasks would require studying how plans guide reasoning over visual or other non-text inputs and how the aggregator evaluates the resulting candidates.

MPPA uses additional inference-time computation relative to single-sample decoding, which increases energy use. Although evaluated on mathematical and scientific reasoning, it may still produce incorrect answers and should not be used in high-stakes settings without domain-specific validation.

%% file: sections/appendix.tex
\clearpage
\section{Additional Methodology Details}
\label{app:additional_methodology}

\subsection{Supervision Construction}
\label{app:supervision_construction}
\label{app:teacher_plan_construction}

We use gold subtask answers as references when available. For synthetic composite problems, GPT-5.6 Sol provides each subtask's reference answer after substituting the correct prerequisite values. We check that intermediate answers are real numbers and that the terminal answer is equivalent to the complete problem's answer using the MATH grader. GPT-5.6 Terra is then asked to solve the complete problem and instantiated subtasks, and verify that every question is well-posed and every proposed answer is correct; only examples passing all checks are retained (\Cref{app:composite_tasks}). For each training subtask, we instantiate its prerequisites with reference answers; subtasks without prerequisites use their original questions directly. We run the frozen executor eight times without a plan, obtaining a correct count $b_0$. When $b_0=8$, we retain one plan generated directly by GPT-5.6 Sol without search. Otherwise, GPT-5.6 Sol proposes a plan of at most 200 tokens, which undergoes rule-based and GPT-5.6 Terra checks for answer leakage, with at most two rewrites if rejected. We execute an approved plan eight times and retain it without search if its correct count is at least $b_0+2$ or seven. The remaining subtasks undergo feedback-guided search: GPT-5.6 Sol revises plans based on the executor's solutions for at most five rounds, stopping after two consecutive rounds without improvement in the best execution score. Rejected plans are repaired within the same round.

Only subtasks entering search undergo independent re-evaluation. We select the three distinct plans with the highest search scores, execute each sixteen times using fresh random seeds, and retain those whose accuracy is at least $b_0/8$, ordered by re-evaluation accuracy. For a subtask with $L_j$ usable plans, we construct $L_j$ SFT examples with budgets $K=1,\ldots,L_j$. Each input contains the instantiated subtask and a planner prompt requesting $K$ distinct approaches; its target contains the first $K$ usable plans. Subtasks retained without search contribute one $K=1$ example, while those without usable plans are excluded.

\subsection{Optimization Details}
\label{app:optimization_details}

For each subtask input $\tilde q_j$, we sample one budget $K$ and generate $G$ plan collections using the same prompt. Let $S_j^{(g)}$ denote the $g$-th sampled collection. We normalize rewards by subtracting the group mean, without dividing by the standard deviation:
\begin{equation}
    A_{j,g}=R(S_j^{(g)})-\frac{1}{G}\sum_{g'=1}^{G}R(S_j^{(g')}).
    \label{eq:planner_group_advantage}
\end{equation}
This preserves the reward scale and avoids amplifying low-variance groups. All output tokens of a plan collection, including the end token, share its advantage. We perform one on-policy GRPO update per batch, summing advantage-weighted token log-likelihoods and dividing by the total number of output tokens. The importance ratio is one, so no clipping is needed. Equal-reward groups have zero advantages and are not resampled. No KL or entropy term is added to the planner loss. Collections that cannot be parsed, are truncated, or fail format or answer-leakage checks receive zero reward and remain in their groups.

Each aggregator training list contains $K$ candidate solutions for the same subtask input. We average pointwise loss over all candidates, including individual candidates replayed from stage-one training, and average listwise loss separately over mixed lists. Individual replay candidates contribute only to the pointwise term. Candidate and mixed-list counts are aggregated across devices before normalization. When no mixed lists are present, the listwise term is zero; the batch is retained and $\eta$ is unchanged.

\subsection{Training and Inference Procedures}
\label{app:training_and_inference}
\label{app:training}
\label{app:inference}

\Cref{alg:mppa_training} reuses the collected candidates for both adapter updates.
In \Cref{alg:mppa_inference}, $\operatorname{Pred}(j)$ denotes the prerequisite subtasks of $q_j$.
The ready set $\mathcal R$ supports branching and merging dependencies; independent
subtasks can run concurrently. \textsc{ComposeTaskAnswer} combines selected subtask
outputs according to the task's required output. All model parameters are fixed
during inference, and each scoring input contains one complete candidate solution.

\input{sections/training_algorithm.tex}
\input{sections/inference_algorithm.tex}

\section{Datasets}
\label{app:datasets}

\paragraph{Training data.}
We construct a shared pool of 18,319 problems spanning mathematics and science. The mathematics portion includes 6,243 level 4--5 problems from the MATH500 training set~\citep{hendrycks2021measuring,lightman2023let} and an AIME-oriented subset of 4,802 problems, comprising 910 historical AIME problems~\citep{AIME2024} from 1983--2023 and 3,892 Omni-MATH problems~\citep{gao2024omnimath}.
The science portion contains 5,000 GPQA-oriented problems and 2,274 MMLU-Pro problems~\citep{wang2024mmlupro}. The GPQA subset combines 343 non-Diamond GPQA questions~\citep{gpqa} with 4,657 medium- and high-difficulty SuperGPQA questions~\citep{map2025supergpqa} in physics, chemistry, and biology, each with four answer options. The MMLU-Pro subset contains 985 physics, 756 chemistry, and 533 biology questions, retaining their original ten-option format.
For MMLU-Pro, we exclude the questions reserved for MMLU-Pro-sci500 evaluation and remove questions that are duplicates of, or highly similar to, questions in the GPQA training subset or MMLU-Pro-sci500.

\paragraph{Composite tasks with provided subtasks.}
\label{app:composite_tasks}
CompMATH contains 100 problems for each subtask count $N\in\{2,3,5\}$, constructed from MATH500 seed questions~\citep{lightman2023let} without reusing a seed within the same $N$. Each dependency graph has one terminal node and in-degree at most two. For $N=2$, all 100 problems use a chain ($1\rightarrow2$). For $N=3$ setting, 51 problems use a fan-in graph ($1,2\rightarrow3$), and 49 use a triangular graph ($1\rightarrow2$, $1\rightarrow3$, and $2\rightarrow3$). For $N=5$ setting, 27 problems use converging chains ($1\rightarrow2$, $3\rightarrow4$, and $2,4\rightarrow5$), 27 use a fan-in tree ($1,2\rightarrow4$ and $3,4\rightarrow5$), 24 use randomly layered DAGs with example-specific edges, and 22 use a diamond graph ($1\rightarrow\{2,3\}$, $\{2,3\}\rightarrow4$, and $4\rightarrow5$). Pure chains are excluded for $N=5$.

We give GPT-5.6 Sol the seed question, its answer, and the prescribed graph. We ask for a new problem with subtasks of comparable topic, technique, and difficulty that does not copy the seed. Downstream subtasks must genuinely use upstream results, rather than merely substitute an earlier answer into an otherwise unrelated question. Intermediate answers must be exact real numbers, preferably integers. GPT-5.6 Sol outputs a natural complete problem that asks only for the final answer, together with a decomposition into standalone subtask templates. Each template uses placeholders such as \texttt{\{s1\}} and \texttt{\{s2\}} for its prerequisite answers and includes a reference answer. Downstream reference answers are computed after substituting the correct prerequisite values.

We discard an example if its placeholders do not match the prescribed graph exactly. We also automatically check that all intermediate reference answers are real numbers and that the terminal subtask's answer is equivalent to the complete problem's answer using the MATH grader. We then make one GPT-5.6 Terra call per remaining example. GPT-5.6 Terra receives the complete problem and proposed final answer, along with every subtask instantiated using the reference prerequisite answers and its proposed answer. It is asked to solve the problems and assess whether each question is unambiguous with a unique answer, whether the proposed answers are correct, and whether the decomposition matches the complete problem. We retain an example only if every judgment is affirmative.

\begin{figure}[t]
    \centering
    \includegraphics[width=\linewidth]{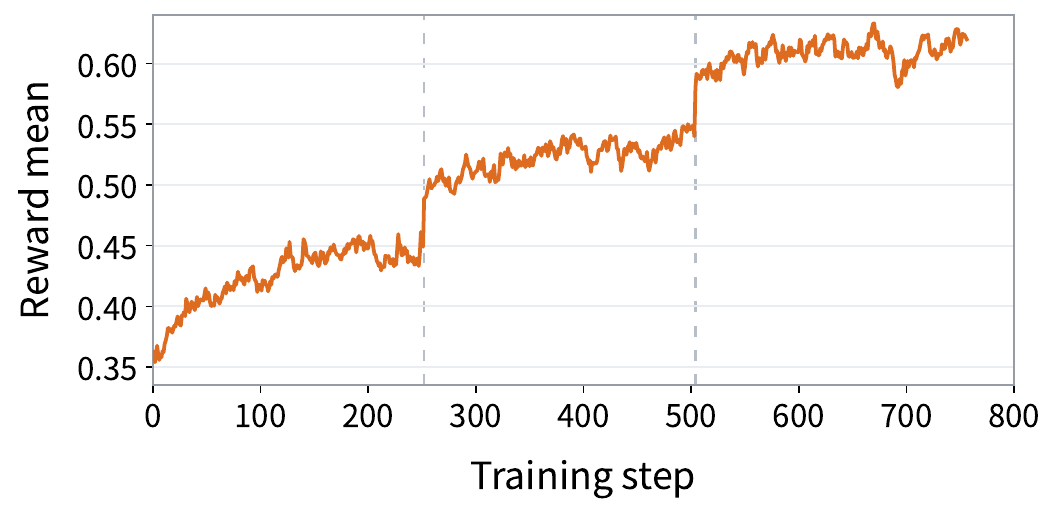}
    \caption{Planner GRPO training curve for Qwen2.5-7B-Instruct, showing batch mean reward over three alternating optimization rounds.}
    \label{fig:grpo_training_curve}
\end{figure}

\paragraph{Dataset examples.}
\label{app:dataset_examples}
\label{app:compmath_examples}

The boxes below show two examples from each evaluation dataset and two CompMATH examples for each $N\in\{2,3,5\}$. The GPQA examples reproduce the Genetics and Astrophysics questions in Table~1 of \citet{gpqa}, with answers from the official dataset and option order preserved from the paper. Each CompMATH example includes the complete problem, dependency graph, subtask templates, and reference answers; placeholders such as \texttt{\{s1\}} denote upstream answers.

\begin{figure*}[!t]
\centering
\begin{mppaprompt}{MATH500 Examples}
\scriptsize
\setlength{\parskip}{0pt}
\textbf{1. \texttt{algebra/1529}}
\par
If $t(x) = 3-g(x)$ and $g(x) = \sqrt{x}$, then what is $t(g(16))$?
\par
\textbf{Answer:} 1
\par\vspace{\baselineskip}
\textbf{2. \texttt{number\_theory/156}}
\par
Let $n$ be a positive integer. What is the greatest possible value of $\gcd(n + 7, 2n + 1)$?
\par
\textbf{Answer:} 13
\end{mppaprompt}
\end{figure*}

\begin{figure*}[!t]
\centering
\begin{mppaprompt}{AIME24 Examples}
\scriptsize
\setlength{\parskip}{0pt}
\textbf{1. \texttt{aime24-84}}
\par
Let $x, y$ and $z$ be positive real numbers that satisfy the following system of equations:
\[
\log_2\left(\frac{x}{yz}\right) = \frac12,\qquad
\log_2\left(\frac{y}{xz}\right) = \frac13,\qquad
\log_2\left(\frac{z}{xy}\right) = \frac14.
\]
Then the value of $\left|\log_2(x^4y^3z^2)\right|$ is $\frac{m}{n}$ where $m$ and $n$ are relatively prime positive integers. Find $m+n$.
\par
\textbf{Answer:} 033
\par\vspace{\baselineskip}
\textbf{2. \texttt{aime24-71}}
\par
Consider the paths of length 16 that follow the lines from the lower left corner to the upper right corner on an $8\times 8$ grid. Find the number of such paths that change direction exactly four times, as in the examples shown below.
\par
\textbf{Answer:} 294
\par
\textit{The original problem includes an illustration, which is absent from the dataset.}
\end{mppaprompt}
\end{figure*}

\begin{figure*}[!t]
\centering
\begin{mppaprompt}{AIME25 Examples}
\scriptsize
\setlength{\parskip}{0pt}
\textbf{1. \texttt{aime25-30}}
\par
There are exactly three positive real numbers $k$ such that the function $f(x) = \frac{(x - 18)(x - 72)(x - 98)(x - k)}{x}$ defined over the positive real numbers achieves its minimum value at exactly two positive real numbers $x$. Find the sum of these three values of $k$.
\par
\textbf{Answer:} 240
\par\vspace{\baselineskip}
\textbf{2. \texttt{aime25-28}}
\par
Let $x_1, x_2, x_3, \ldots$ be a sequence of rational numbers defined by $x_1 = \frac{25}{11}$ and $x_{k+1} = \frac13\left(x_k + \frac{1}{x_k} - 1\right)$ for all $k \ge 1$. Then $x_{2025}$ can be expressed as $\frac{m}{n}$, where $m$ and $n$ are relatively prime positive integers. Find the remainder when $m + n$ is divided by 1000.
\par
\textbf{Answer:} 248
\end{mppaprompt}
\end{figure*}

\begin{figure*}[!t]
\centering
\begin{mppaprompt}{GPQA Examples}
\scriptsize
\setlength{\parskip}{0pt}
\textbf{1. Genetics} (\citealp{gpqa}, Table~1)
\par
If a sperm from species A is injected into an egg from species B and both species have the same number of chromosomes, what would be the main cause of the resulting zygote mortality?
\par\vspace{0.5\baselineskip}
(A) Species specific zona pellucida proteins on the egg cannot bind sperms from a different species.
\par
(B) Epistatic interactions between the genes of different species
\par
(C) Chromosomal incompatibilities will cause failure of meiosis leading to death of zygote.
\par
(D) Chromosomal recombination will not occur in different species.
\par
\textbf{Answer:} B
\par\vspace{\baselineskip}
\textbf{2. Astrophysics} (\citealp{gpqa}, Table~1)
\par
Astronomers are studying a star with a Teff of approximately 6000 K. They are interested in spectroscopically determining the surface gravity of the star using spectral lines (EW $<$ 100 mA) of two chemical elements, El1 and El2. Given the atmospheric temperature of the star, El1 is mostly in the neutral phase, while El2 is mostly ionized. Which lines are the most sensitive to surface gravity for the astronomers to consider?
\par\vspace{0.5\baselineskip}
(A) El2 I (neutral)
\par
(B) El1 II (singly ionized)
\par
(C) El2 II (singly ionized)
\par
(D) El1 I (neutral)
\par
\textbf{Answer:} B
\end{mppaprompt}
\end{figure*}

\begin{figure*}[!t]
\centering
\begin{mppaprompt}{MMLU-Pro-sci500 Examples}
\scriptsize
\setlength{\parskip}{0pt}
\textbf{1. \texttt{MMLUPRO:9580} (Physics)}
\par
What is the focal length of a combination of two thin convex lenses, each of focal length $f$, placed apart at a distance of $2f/3$?
\par\vspace{0.5\baselineskip}
(A) $f/3$ \quad (B) $f/2$ \quad (C) $5f/3$ \quad (D) $6f/5$ \quad (E) $3f/4$ \quad
(F) $4f/3$ \quad (G) $2f/3$ \quad (H) $f/4$ \quad (I) $2f$ \quad (J) $7f/6$
\par
\textbf{Answer:} E
\par\vspace{\baselineskip}
\textbf{2. \texttt{MMLUPRO:2998} (Biology)}
\par
The amino acid ornithine is regenerated during each turn of the urea cycle in a similar manner to which of the following participants in the citric acid cycle?
\par\vspace{0.5\baselineskip}
(A) Alpha-ketoglutarate \quad (B) Glutamate \quad (C) Acetyl CoA \quad
(D) Oxaloacetate \quad (E) Citrate \quad (F) Aspartate \quad
(G) Fumarate \quad (H) Malate \quad (I) Succinate \quad (J) Isocitrate
\par
\textbf{Answer:} D
\end{mppaprompt}
\end{figure*}

\begin{figure*}[!t]
\centering
\begin{mppaprompt}{CompMATH Examples: $N=2$}
\scriptsize
\setlength{\parskip}{0pt}
\textbf{Example 1}
\par
\textbf{Graph:} chain ($1\rightarrow2$).
\par\vspace{0.5\baselineskip}
\textbf{Problem.} The expression $\dfrac{2a+6}{3}-\dfrac{a+5}{2}$, written as a single fraction, equals zero exactly when $a=c$. Using this value of $c$, write $\frac{c+4a}{7}-\frac{2-ca}{5}$ as a single fraction.
\par
\textbf{Answer:} $\frac{1+41a}{35}$.
\par\vspace{\baselineskip}
\textbf{Subtask 1 (no prerequisites).} Write $\dfrac{2a+6}{3}-\dfrac{a+5}{2}$ as a single fraction, and find the value of $a$ that makes it equal to $0$.
\par
\textbf{Answer:} 3.
\par\vspace{0.5\baselineskip}
\textbf{Subtask 2 (depends on 1).} Write the expression $\frac{\texttt{\{s1\}}+4a}{7}-\frac{2-\texttt{\{s1\}}a}{5}$ as a single fraction.
\par
\textbf{Answer:} $\frac{1+41a}{35}$.
\par\vspace{\baselineskip}
\textbf{Example 2}
\par
\textbf{Graph:} chain ($1\rightarrow2$).
\par\vspace{0.5\baselineskip}
\textbf{Problem.} At both Lincoln and Jefferson Middle School, elections are held for president, vice president, and secretary. At each school, every office is contested by exactly two candidates---one from the Eagle Party and one from the Hawk Party---and each candidate has a 50\% chance of winning, independently of all other races. The two schools' elections are independent of each other. What is the probability that at least one of the two schools elects an all-Eagle slate, meaning the Eagle Party candidate wins all three offices at that school?
\par
\textbf{Answer:} $\frac{15}{64}$.
\par\vspace{\baselineskip}
\textbf{Subtask 1 (no prerequisites).} At Lincoln Middle School, elections are held for president, vice president, and secretary. Each office is contested by exactly two candidates---one from the Eagle Party and one from the Hawk Party---and each candidate has a 50\% chance of winning, independently of the other races. What is the probability that the Eagle Party candidate wins all three offices? (Express your answer as a fraction.)
\par
\textbf{Answer:} $\frac18$.
\par\vspace{0.5\baselineskip}
\textbf{Subtask 2 (depends on 1).} Lincoln and Jefferson Middle School hold their elections independently of each other, and at each school the probability that the Eagle Party candidate wins all three offices is \texttt{\{s1\}}. What is the probability that at least one of the two schools elects an all-Eagle slate? (Express your answer as a fraction.)
\par
\textbf{Answer:} $\frac{15}{64}$.
\end{mppaprompt}
\end{figure*}

\begin{figure*}[!t]
\centering
\begin{mppaprompt}{CompMATH Examples: $N=3$}
\scriptsize
\setlength{\parskip}{0pt}
\textbf{Example 1}
\par
\textbf{Graph:} fan-in ($1\rightarrow3$, $2\rightarrow3$).
\par\vspace{0.5\baselineskip}
\textbf{Problem.} The sum of the first $m$ positive odd numbers is $400$, and the sum of the first $k$ positive even numbers is $272$. Find the sum of all odd numbers that lie strictly between $2k$ and $2m$.
\par
\textbf{Answer:} 144.
\par\vspace{\baselineskip}
\textbf{Subtask 1 (no prerequisites).} The sum of the first $m$ positive odd numbers is $400$. What is the value of $m$?
\par
\textbf{Answer:} 20.
\par\vspace{0.5\baselineskip}
\textbf{Subtask 2 (no prerequisites).} The sum of the first $k$ positive even numbers is $272$. What is the value of $k$?
\par
\textbf{Answer:} 16.
\par\vspace{0.5\baselineskip}
\textbf{Subtask 3 (depends on 1 and 2).} Let $m=\texttt{\{s1\}}$ and $k=\texttt{\{s2\}}$. Find the sum of all odd numbers that lie strictly between $2k$ and $2m$.
\par
\textbf{Answer:} 144.
\par\vspace{\baselineskip}
\textbf{Example 2}
\par
\textbf{Graph:} triangle ($1\rightarrow2$, $1\rightarrow3$, $2\rightarrow3$).
\par\vspace{0.5\baselineskip}
\textbf{Problem.} Let $a$ be the unique number with $0<a<180$ such that
\[
\tan32^\circ\tan76^\circ\tan a^\circ
=\tan32^\circ+\tan76^\circ+\tan a^\circ,
\]
and let $b$ be the smallest positive number such that
\[
\tan b^\circ\tan2b^\circ\tan a^\circ
=\tan b^\circ+\tan2b^\circ+\tan a^\circ.
\]
Find the smallest positive number $c$ such that
\[
\tan c^\circ\tan(c+b)^\circ\tan(c+a)^\circ
=\tan c^\circ+\tan(c+b)^\circ+\tan(c+a)^\circ.
\]
\textbf{Answer:} 24.
\par\vspace{\baselineskip}
\textbf{Subtask 1 (no prerequisites).} Find the unique $x$ with $0<x<180$ such that
$\tan32^\circ\tan76^\circ\tan x^\circ=\tan32^\circ+\tan76^\circ+\tan x^\circ$.
\par
\textbf{Answer:} 72.
\par\vspace{0.5\baselineskip}
\textbf{Subtask 2 (depends on 1).} Find the smallest positive number $y$ such that
$\tan y^\circ\tan2y^\circ\tan\texttt{\{s1\}}^\circ=\tan y^\circ+\tan2y^\circ+\tan\texttt{\{s1\}}^\circ$.
\par
\textbf{Answer:} 36.
\par\vspace{0.5\baselineskip}
\textbf{Subtask 3 (depends on 1 and 2).} Find the smallest positive number $z$ such that
$\tan z^\circ\tan(z+\texttt{\{s2\}})^\circ\tan(z+\texttt{\{s1\}})^\circ
=\tan z^\circ+\tan(z+\texttt{\{s2\}})^\circ+\tan(z+\texttt{\{s1\}})^\circ$.
\par
\textbf{Answer:} 24.
\end{mppaprompt}
\end{figure*}

\begin{figure*}[!t]
\centering
\begin{mppaprompt}{CompMATH Examples: $N=5$}
\scriptsize
\setlength{\parskip}{0pt}
\textbf{Example 1}
\par
\textbf{Graph:} diamond ($1\rightarrow\{2,3\}$, $\{2,3\}\rightarrow4$, $4\rightarrow5$).
\par\vspace{0.5\baselineskip}
\textbf{Problem.} A convex polygon $P$ has the property that the sum of all but one of its interior angles is $2570^\circ$. Let $d$ be the number of diagonals of $P$, and let $m$ be the number of sides of the regular polygon each of whose interior angles measures one eighteenth of the sum of the interior angles of $P$. Now let $N$ be the number of sides of a convex polygon in which the sum of all but one of the interior angles is $d\cdot m$ degrees. Finally, the interior angles of a convex polygon $Q$ with $N$ sides form an arithmetic progression whose smallest term is $126^\circ$. Find the largest interior angle of $Q$.
\par
\textbf{Answer:} $162^\circ$.
\par\vspace{\baselineskip}
\textbf{Subtask 1 (no prerequisites).} A convex polygon has the property that the sum of all but one of its interior angles is $2570^\circ$. How many sides does the polygon have?
\par
\textbf{Answer:} 17.
\par\vspace{0.5\baselineskip}
\textbf{Subtask 2 (depends on 1).} How many diagonals does a convex polygon with \texttt{\{s1\}} sides have?
\par
\textbf{Answer:} 119.
\par\vspace{0.5\baselineskip}
\textbf{Subtask 3 (depends on 1).} Let $S$ be the sum of the interior angles of a convex polygon with \texttt{\{s1\}} sides. A regular polygon has each of its interior angles equal to $\frac{S}{18}$. How many sides does this regular polygon have?
\par
\textbf{Answer:} 12.
\par\vspace{0.5\baselineskip}
\textbf{Subtask 4 (depends on 2 and 3).} In a convex polygon, the sum of all but one of its interior angles is $\texttt{\{s2\}}\times\texttt{\{s3\}}$ degrees. How many sides does this polygon have?
\par
\textbf{Answer:} 10.
\par\vspace{0.5\baselineskip}
\textbf{Subtask 5 (depends on 4).} The interior angles of a convex polygon with \texttt{\{s4\}} sides form an arithmetic progression whose smallest term is $126^\circ$. What is the measure of its largest interior angle?
\par
\textbf{Answer:} $162^\circ$.
\par\vspace{\baselineskip}
\textbf{Example 2}
\par
\textbf{Graph:} two merging chains ($1\rightarrow2$, $3\rightarrow4$, $\{2,4\}\rightarrow5$).
\par\vspace{0.5\baselineskip}
\textbf{Problem.} A school supply closet holds 80 boxes. Thirty of the boxes contain pencils, 26 contain pens, and 35 contain erasers. It is known that 44 of the boxes contain pencils or pens but not both, that 14 boxes contain both pencils and erasers, that 11 boxes contain both pens and erasers, and that 15 boxes contain erasers together with exactly one of pencils and pens. How many of the boxes contain none of the three items?
\par
\textbf{Answer:} 15.
\par\vspace{\baselineskip}
\textbf{Subtask 1 (no prerequisites).} In a supply closet, 30 boxes contain pencils and 26 boxes contain pens. If 44 of the boxes contain pencils or pens but not both, how many boxes contain both pencils and pens?
\par
\textbf{Answer:} 6.
\par\vspace{0.5\baselineskip}
\textbf{Subtask 2 (depends on 1).} In a supply closet, 30 boxes contain pencils, 26 boxes contain pens, and \texttt{\{s1\}} boxes contain both pencils and pens. How many boxes contain at least one of pencils and pens?
\par
\textbf{Answer:} 50.
\par\vspace{0.5\baselineskip}
\textbf{Subtask 3 (no prerequisites).} In a supply closet, 14 boxes contain both pencils and erasers and 11 boxes contain both pens and erasers. If 15 boxes contain erasers together with exactly one of pencils and pens, how many boxes contain all three of pencils, pens, and erasers?
\par
\textbf{Answer:} 5.
\par\vspace{0.5\baselineskip}
\textbf{Subtask 4 (depends on 3).} In a supply closet, 14 boxes contain both pencils and erasers, 11 boxes contain both pens and erasers, and \texttt{\{s3\}} boxes contain all three of pencils, pens, and erasers. How many boxes contain erasers together with at least one of pencils and pens?
\par
\textbf{Answer:} 20.
\par\vspace{0.5\baselineskip}
\textbf{Subtask 5 (depends on 2 and 4).} A supply closet holds 80 boxes. Of these, \texttt{\{s2\}} boxes contain at least one of pencils and pens, 35 boxes contain erasers, and \texttt{\{s4\}} boxes contain erasers together with at least one of pencils and pens. How many boxes contain none of pencils, pens, and erasers?
\par
\textbf{Answer:} 15.
\end{mppaprompt}
\end{figure*}

\section{Implementation Details}
\label{app:implementation_details}

\subsection{Model and Training Configurations}
\label{app:training_configurations}

\paragraph{Planner SFT.}
We use LoRA with rank $r=32$, $\alpha=64$, dropout $0.05$, and bias adaptation disabled (\texttt{bias=none}). For Qwen and Gemma, we target seven linear projections: \texttt{q\_proj}, \texttt{k\_proj}, \texttt{v\_proj}, \texttt{o\_proj}, \texttt{gate\_proj}, \texttt{up\_proj}, and \texttt{down\_proj}. For Phi-4, we target \texttt{qkv\_proj}, \texttt{o\_proj}, \texttt{gate\_up\_proj}, and \texttt{down\_proj}, which cover the corresponding attention and MLP projections in fused form. The backbone is frozen and stored in bfloat16, while LoRA parameters and AdamW states use float32. We train for two epochs using AdamW with a constant learning rate of $10^{-4}$, zero weight decay, and a gradient clipping threshold of $1.0$.

\paragraph{Planner GRPO.}
Starting from the Planner SFT adapter, we train with GRPO for $T=3$ alternating rounds at a learning rate of $5\times10^{-5}$. Before GRPO training, we prescreen subtask inputs using eight sampled solutions and exclude those for which all eight answers are correct. Each rollout batch contains 64 subtask inputs, with $G=8$ plan collections sampled per input. We perform one update per batch (\texttt{ppo\_epochs=1}), with the mini-batch spanning the entire rollout batch. We use AdamW with zero weight decay and a gradient clipping threshold of $1.0$. Both KL and entropy coefficients are zero; \texttt{low\_var\_kl} is recorded for monitoring and does not enter the loss. Rollouts use temperature $0.7$, top-$p$ $0.8$, top-$k$ $20$, and repetition penalty $1.05$, with a maximum output length of 1,024 tokens per complete plan collection. Rewards follow \Cref{eq:planner_reward} with $\lambda=0.5$; collections that cannot be parsed, are truncated, or fail format or answer-leakage checks receive zero reward.
A three-round GRPO run takes approximately 50 hours on a single A100 40GB node.
\Cref{fig:grpo_training_curve} shows the GRPO training curve.

\paragraph{Aggregator SFT.}
The aggregator uses the same LoRA rank, $\alpha$, target modules, and bias setting as Planner SFT. Both stages keep the bfloat16 backbone frozen and use float32 LoRA parameters and AdamW states. We use AdamW with zero weight decay and a gradient clipping threshold of $1.0$. Stage one trains for two epochs with a learning rate of $10^{-4}$ and dropout $0.05$. Stage two continues from the stage-one adapter with a learning rate of $3\times10^{-5}$, sets dropout to zero, and uses a listwise loss weight of $\eta=1$. In each alternating round, we first train the planner for one epoch, then train the aggregator for one pass over its training data, and finally evaluate the updated pair on the validation set.

\subsection{Prompts and Output Processing}
\label{app:prompts}

The templates below are taken directly from our implementation. Each module receives a single user message with no system prompt, wrapped using the backbone's chat template.

\paragraph{Planner.}
SFT and inference use the same template. The \promptfield{kind} field is ``math problem'' for MATH500, AIME24/25, and synthetic subtasks, and ``multiple-choice science question'' for GPQA and MMLU-Pro. We substitute the requested budget for \promptfield{K} and use ``es'' for \promptfield{es} when $K>1$, otherwise an empty string. The SFT target and generated output contain sketches enclosed in numbered \texttt{<sketch>} tags.

\paragraph{Plan auditing during RL.}
We audit sketches sequentially before execution. If any sketch fails either audit, the entire plan collection receives zero reward and is not executed. We first apply \texttt{rule\_audit}, rejecting an empty sketch, a sketch exceeding 200 tokens, or a sketch containing at least two occurrences of ``step $N$,'' indicating a step-by-step derivation. For multiple-choice questions, we also reject sketches containing option letters or substantial quotations of the correct option. For mathematical questions, we reject sketches containing the normalized reference-answer string or a numerical value equal to the reference answer.

Only sketches passing the rule audit are sent to GPT-5.6 Terra, together with the question and reference answer. GPT-5.6 Terra returns a JSON object with fields \texttt{leaks\_answer}, \texttt{rule}, and \texttt{reason}, checking whether the sketch (a) states the answer or correct option; (b) supplies a decisive intermediate result or structural conclusion; (c) gives a step-by-step derivation or a formula leaving only arithmetic; (d) singles out one candidate answer; or (e) excludes or implies an option through a pitfall warning. Methods, principles, objectives to establish, relevant theorems, checks to perform, and pitfalls at the method level are explicitly permitted.

\begin{figure*}[t]
\centering
\begin{mppaprompt}{Planner Prompt Template}
\scriptsize
\setlength{\parskip}{0pt}
Write \promptfield{K} different short sketch\promptfield{es} of the approach for the \promptfield{kind} below. Each sketch will be given, on its own and together with the problem, to an executor model, which then solves the problem itself. Each sketch: at most 110 words; the rough approach only (method, key observation, relevant theorems, main pitfall); no answer, no derived intermediate results, no step-by-step derivation; sketches should follow different routes, best first.
\par\vspace{\baselineskip}
Problem:
\par
\promptfield{problem}
\par\vspace{\baselineskip}
\textbf{Output format:}
\par
\texttt{<sketch id="1">}\ldots\texttt{</sketch>}
\par
\texttt{<sketch id="2">}\ldots\texttt{</sketch>}
\par
\ldots
\end{mppaprompt}
\end{figure*}

\paragraph{Executor.}
The executor receives either a sketch with the problem or the problem alone. During subtask inference, \promptfield{problem} is the subtask question with placeholders such as \texttt{\{sj\}} replaced by the selected upstream answers. The dataset-specific \promptfield{suffix} is:
\begin{itemize}[leftmargin=*,nosep]
    \item MATH500, AIME24/25, and synthetic subtasks: ``Reason step by step, and put only the final answer inside \texttt{\textbackslash boxed\{...\}}.''
    \item GPQA: ``Reason step by step, and put only the letter of the correct option (A, B, C, or D) inside \texttt{\textbackslash boxed\{...\}}.''
    \item MMLU-Pro: ``Reason step by step, and put only the letter of the correct option (one of A to J) inside \texttt{\textbackslash boxed\{...\}}.''
\end{itemize}

\begin{figure*}[!t]
\centering
\begin{mppaprompt}{Executor Prompt Templates}
\scriptsize
\setlength{\parskip}{0pt}
\textbf{With a plan:}
\par\vspace{\baselineskip}
Hint for the problem below (a suggested approach; it may be imperfect, deviate from it if it is wrong):
\par
\promptfield{sketch}
\par\vspace{\baselineskip}
Problem:
\par
\promptfield{problem}
\par\vspace{\baselineskip}
The approach in the hint is only a hint, not a solution. Work out the complete solution yourself, step by step, before giving the final answer. \promptfield{suffix}
\par\vspace{\baselineskip}
\textbf{Without a plan:}
\par\vspace{\baselineskip}
\promptfield{problem}
\par\vspace{\baselineskip}
\promptfield{suffix}
\end{mppaprompt}
\end{figure*}
\begin{figure*}[!t]
\centering
\begin{mppaprompt}{Aggregator Prompt Template}
\scriptsize
\setlength{\parskip}{0pt}
Here is a \promptfield{kind} and a candidate solution. Is the candidate's final answer correct? Answer Yes or No.
\par\vspace{\baselineskip}
Problem:
\par
\promptfield{problem}
\par\vspace{\baselineskip}
Candidate solution:
\par
\promptfield{solution}
\end{mppaprompt}
\end{figure*}

\paragraph{Aggregator.}
The aggregator receives the original question for the candidate and the executor's complete output as \promptfield{solution}, using the same \promptfield{kind} values as the planner. For subtasks, the question contains the instantiated upstream answers. The SFT target is one token, \texttt{Yes} or \texttt{No}. At inference, we generate one token at temperature zero and request the top-20 token log-probabilities, scoring each candidate by $\log P(\texttt{Yes})-\log P(\texttt{No})$. We select the candidate with the highest score.

\begin{table*}[t]
\centering
\small
\caption{Candidate coverage and selection accuracy (\%) for Qwen2.5-7B-Instruct with $K=3$ candidates. Each cell reports agg@K / pass@K, where agg@K uses the same learned aggregator.}
\label{tab:candidate_coverage_selection}
\setlength{\tabcolsep}{5pt}
\resizebox{0.75\linewidth}{!}{%
\begin{tabular}{@{}lcccccc@{}}
\toprule
\multirow[c]{2}{*}{Generation} & \multicolumn{3}{c}{Mathematics} & \multicolumn{2}{c}{Science} & \multirow[c]{2}{*}{Avg} \\
\cmidrule(lr){2-4}\cmidrule(lr){5-6}
 & MATH500 & AIME24 & AIME25 & GPQA & MMLU-Pro & \\
\midrule
Base & 83.6 / 84.2 & 17.8 / 18.9 & 13.3 / 14.4 & 54.0 / 54.0 & 78.3 / 78.3 & 49.4 / 50.0 \\
Single plan & 88.0 / 89.6 & 18.9 / 24.4 & 16.7 / 18.9 & 67.2 / 68.2 & 80.1 / 80.2 & 54.2 / 56.3 \\
Independent $K$ plans & 87.2 / 89.3 & 20.0 / 23.3 & 13.3 / 15.6 & 69.0 / 70.2 & 81.4 / 83.5 & 54.2 / 56.4 \\
\rowcolor[gray]{0.94}
Joint $K$ plans & \textbf{89.2 / 92.3} & \textbf{22.2 / 26.7} & \textbf{24.4 / 24.4} & \textbf{72.2 / 73.2} & \textbf{84.0 / 86.0} & \textbf{58.4 / 60.5} \\
\bottomrule
\end{tabular}%
}
\end{table*}

\begin{figure*}[t]
\centering
\includegraphics[width=0.8\textwidth]{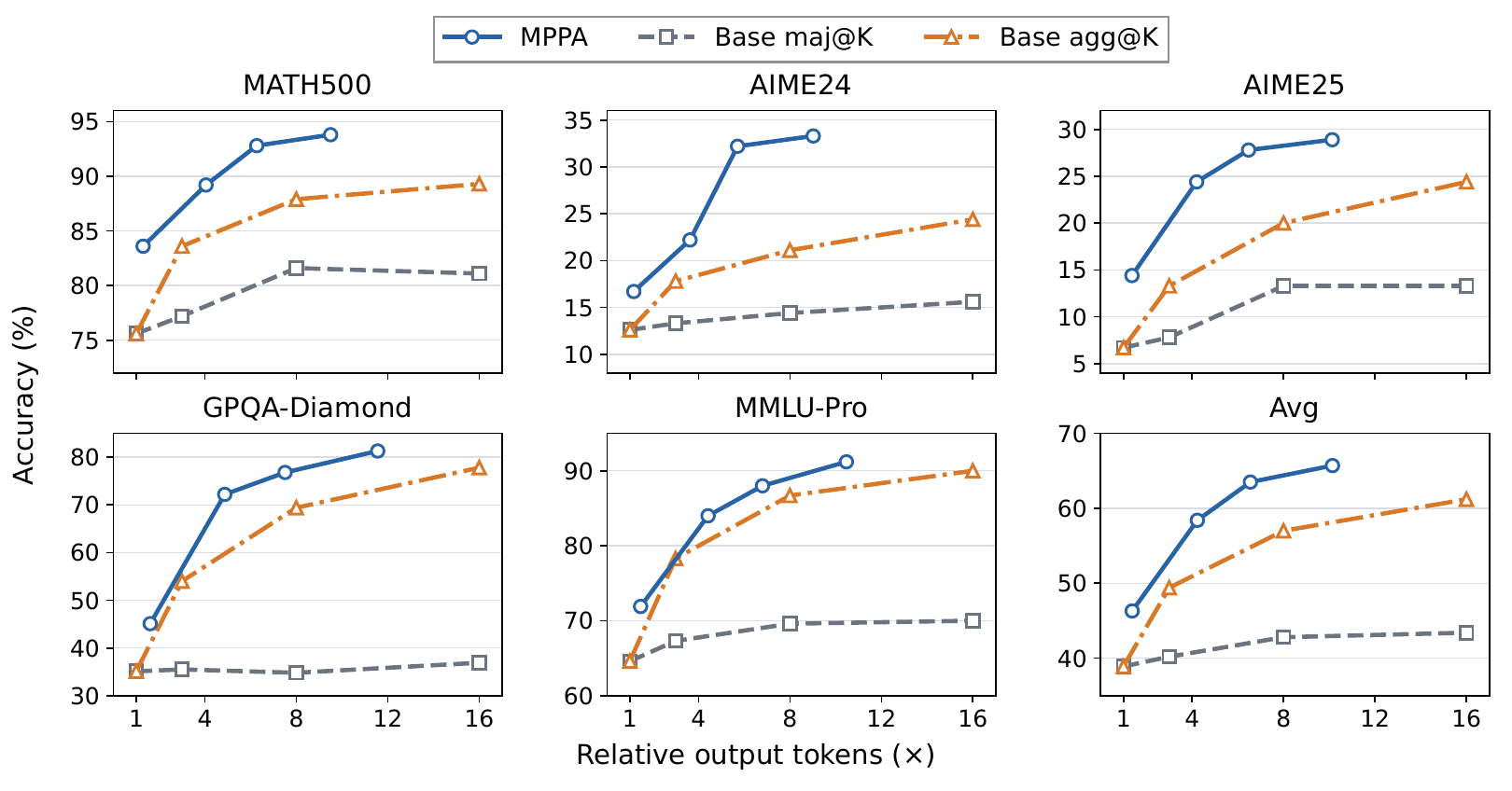}
\caption{Accuracy--output-token trade-offs for Qwen2.5-7B-Instruct on each dataset and on average. MPPA uses $K=1,3,5,8$ plans; Base maj@K and Base agg@K use $K=1,3,8,16$ directly sampled solutions, with agg@K using the same learned aggregator. Token usage is relative to one directly sampled solution on each dataset; the Avg panel averages accuracies and relative token usage across datasets.}
\label{fig:dataset_accuracy_cost}
\end{figure*}

\begin{table*}[t]
\centering
\footnotesize
\caption{Accuracy (\%) with confidence intervals for the principal sampling and aggregation settings. Each cell reports the mean followed by the interval bounds; means match \Cref{tab:main_results}. GPQA and MMLU-Pro denote GPQA Diamond and MMLU-Pro-sci500.}
\label{tab:main_results_intervals}
\setlength{\tabcolsep}{4pt}
\renewcommand{\arraystretch}{1.15}
\resizebox{0.65\linewidth}{!}{%
\begin{tabular}{@{}lccccc@{}}
\toprule
Method & MATH500 & AIME24 & AIME25 & GPQA & MMLU-Pro \\
\midrule
\multicolumn{6}{c}{Qwen2.5-7B-Instruct} \\
\midrule
Base & 75.6 {\scriptsize [71.9, 78.0]} & 12.6 {\scriptsize [3.6, 20.6]} & 6.7 {\scriptsize [2.1, 12.2]} & 35.1 {\scriptsize [30.3, 39.5]} & 64.6 {\scriptsize [61.1, 67.6]} \\
Base maj@8 & 81.6 {\scriptsize [77.1, 83.5]} & 14.4 {\scriptsize [4.9, 28.9]} & 13.3 {\scriptsize [3.1, 23.1]} & 34.8 {\scriptsize [31.0, 42.2]} & 69.6 {\scriptsize [66.0, 73.3]} \\
Base agg@8 & 87.9 {\scriptsize [85.2, 90.4]} & 21.1 {\scriptsize [8.8, 35.1]} & 20.0 {\scriptsize [8.1, 33.5]} & 69.4 {\scriptsize [64.5, 75.5]} & 86.7 {\scriptsize [83.8, 89.0]} \\
\rowcolor[gray]{0.94}
MPPA@1 & 83.6 {\scriptsize [80.7, 86.3]} & 16.7 {\scriptsize [7.4, 26.7]} & 14.4 {\scriptsize [5.6, 26.3]} & 45.1 {\scriptsize [39.7, 50.7]} & 71.9 {\scriptsize [68.5, 75.1]} \\
\rowcolor[gray]{0.94}
MPPA@3 & 89.2 {\scriptsize [86.4, 91.8]} & 22.2 {\scriptsize [10.0, 36.7]} & 24.4 {\scriptsize [12.2, 37.8]} & 72.2 {\scriptsize [65.7, 78.3]} & 84.0 {\scriptsize [80.8, 87.2]} \\
\midrule
\multicolumn{6}{c}{Gemma-3-12B-it} \\
\midrule
Base & 85.6 {\scriptsize [82.8, 88.3]} & 21.9 {\scriptsize [11.1, 34.0]} & 18.2 {\scriptsize [7.6, 30.3]} & 39.7 {\scriptsize [34.9, 44.7]} & 66.8 {\scriptsize [63.3, 70.1]} \\
Base maj@8 & 89.2 {\scriptsize [86.4, 91.8]} & 27.8 {\scriptsize [14.4, 43.3]} & 21.1 {\scriptsize [8.9, 34.4]} & 44.9 {\scriptsize [37.9, 52.0]} & 72.6 {\scriptsize [68.6, 76.4]} \\
Base agg@8 & 92.8 {\scriptsize [90.4, 95.0]} & 34.4 {\scriptsize [18.9, 51.1]} & 31.1 {\scriptsize [16.7, 46.7]} & 73.7 {\scriptsize [67.7, 79.8]} & 87.2 {\scriptsize [84.2, 90.0]} \\
\rowcolor[gray]{0.94}
MPPA@1 & 88.3 {\scriptsize [85.9, 90.7]} & 40.0 {\scriptsize [27.0, 54.1]} & 18.9 {\scriptsize [7.4, 31.5]} & 49.3 {\scriptsize [43.8, 54.9]} & 75.8 {\scriptsize [72.7, 78.9]} \\
\rowcolor[gray]{0.94}
MPPA@3 & 93.2 {\scriptsize [90.8, 95.4]} & 52.2 {\scriptsize [36.7, 66.7]} & 31.1 {\scriptsize [17.8, 45.6]} & 70.7 {\scriptsize [64.1, 76.8]} & 83.8 {\scriptsize [80.6, 87.0]} \\
\midrule
\multicolumn{6}{c}{Phi-4 (14B)} \\
\midrule
Base & 80.6 {\scriptsize [77.6, 83.5]} & 16.7 {\scriptsize [6.9, 28.3]} & 18.1 {\scriptsize [7.9, 30.2]} & 54.2 {\scriptsize [48.7, 59.7]} & 79.4 {\scriptsize [76.3, 82.2]} \\
Base maj@8 & 84.4 {\scriptsize [81.2, 87.4]} & 21.1 {\scriptsize [9.0, 35.5]} & 23.3 {\scriptsize [10.7, 37.9]} & 59.1 {\scriptsize [52.0, 65.7]} & 83.6 {\scriptsize [80.2, 86.8]} \\
Base agg@8 & 91.0 {\scriptsize [88.4, 93.4]} & 31.1 {\scriptsize [16.8, 46.8]} & 31.1 {\scriptsize [15.9, 47.3]} & 81.8 {\scriptsize [76.3, 86.9]} & 91.8 {\scriptsize [89.4, 94.0]} \\
\rowcolor[gray]{0.94}
MPPA@1 & 87.2 {\scriptsize [84.6, 89.7]} & 26.7 {\scriptsize [15.9, 38.9]} & 24.4 {\scriptsize [13.0, 37.4]} & 65.0 {\scriptsize [59.6, 70.4]} & 84.1 {\scriptsize [81.3, 86.8]} \\
\rowcolor[gray]{0.94}
MPPA@3 & 92.4 {\scriptsize [90.0, 94.6]} & 40.0 {\scriptsize [25.6, 54.4]} & 31.1 {\scriptsize [17.8, 46.7]} & 82.3 {\scriptsize [76.8, 87.4]} & 90.4 {\scriptsize [87.8, 92.8]} \\
\bottomrule
\end{tabular}%
}
\end{table*}

\subsection{Baseline Implementations}
\label{app:baseline_implementations}

\paragraph{Base \& Majority Voting (maj)~\citep{wang2023selfconsistency}.}
Base reports mean@1 accuracy, computed by averaging the correctness of the eight samples per problem collected for Base maj@8. Base maj@8 uses majority voting, grouping equivalent final answers and selecting the most frequent answer, breaking ties by the earliest occurrence.

\paragraph{Plan-and-Solve Prompting (PS+)~\citep{wang2023planandsolve}.}
We insert the original PS+ instruction between the question and the answer-format requirement, asking the model to understand the problem, extract variables, devise a plan, and execute it. We sample eight solutions and apply the same majority-voting procedure as Base maj@8.

\paragraph{Least-to-Most Prompting (L2M)~\citep{zhou2023leasttomost}.}
Our zero-shot Least-to-Most implementation first asks the model to list at most six subproblems, with the original question as the last subproblem. The model then answers them sequentially within the same conversation. We extract the boxed answer to the last subproblem and apply majority voting over eight such chains.

\paragraph{Universal Self-Consistency (USC)~\citep{chen2023universal}.}
We concatenate the eight Base responses into a selection prompt and ask the same model to choose the most consistent response. This reuses the Base samples and requires one additional selection call per problem. If the selection cannot be parsed, we fall back to majority voting.

\paragraph{Early-Stopping Self-Consistency (ESC)~\citep{li2024earlystopping}.}
We sample in batches of $w$ and stop early when all answers within a batch agree. Otherwise, we collect eight samples and apply majority voting. We use $w=2$, requiring an average of 2.6--6.7 samples per problem across the evaluated settings.

\paragraph{Supervised Fine-Tuning (SFT).}
We use gpt-oss-120b~\citep{openai2025gptoss} as the teacher for complete task solutions, retaining only samples that are not truncated, have correct answers, and contain a nonempty reasoning section. The training target is the teacher's complete output, with reasoning enclosed in \texttt{<think>} and \texttt{</think>} tags, followed by the response:
\begin{lstlisting}[basicstyle=\footnotesize\ttfamily,mathescape=false,frame=none,xleftmargin=0pt,xrightmargin=0pt]
"<think>\nreasoning\n</think>\nresponse"
\end{lstlisting}
We apply LoRA to all linear layers with rank $r=32$, $\alpha=64$, and dropout $0.05$. We train for two epochs using AdamW with a constant learning rate of $10^{-4}$, zero weight decay, and a gradient clipping threshold of $1.0$, evaluating validation loss for each data pool every 50 steps. The maximum sequence length is 8,192 tokens; no training sample is truncated by this limit.

\paragraph{Group Relative Policy Optimization (GRPO)~\citep{shao2024deepseekmath}.}
We train on the same training pool, starting from the base model with a newly initialized LoRA adapter, without SFT initialization. Before GRPO training, we prescreen questions using eight sampled solutions and exclude those for which all eight answers are correct. The reward is one if the final answer is correct and zero otherwise. We use standard GRPO, subtracting the group mean reward and dividing by the group standard deviation to obtain advantages, without dynamic sampling. Each batch contains 64 questions with eight sampled solutions per question. The loss uses \texttt{seq-mean-token-mean} reduction: we average over output tokens within each sequence, then over sequences. Each batch receives one on-policy update, so the probability ratio is one and clipping with threshold $0.2$ is inactive. Both KL and entropy coefficients are zero. We apply LoRA with rank $r=32$ and $\alpha=64$ to all linear layers, using AdamW with a constant learning rate of $10^{-5}$ and a gradient clipping threshold of $1.0$. Rollouts use the same sampling parameters as evaluation, with a maximum output length of 8,192 tokens.

\section{Additional Results}
\label{app:additional_results}

\subsection{Example Outputs}
\label{app:example_outputs}

We show MPPA@3 outputs on MATH500, MMLU-Pro-sci500, and a CompMATH task with three subtasks, using the input templates in \Cref{app:prompts}. For each question or subtask, we report three jointly generated sketches, the complete executor responses, and aggregator scores. In the MATH500 example, the third sketch leads to an incorrect execution. In the science example, the aggregator selects the correct candidate even though the other two candidates agree on an incorrect option. The CompMATH example uses a triangular dependency graph: $1\rightarrow2$, $1\rightarrow3$, and $2\rightarrow3$. The selected answers $AC=12$ and $AD=15$ are substituted into downstream questions. In the final subtask, learned aggregation selects $56/65$ even though two of the three candidates return $12/13$.

\begin{figure*}[!t]
\centering
\begin{mppaprompt}{MATH500: Planner Output}
\scriptsize
\setlength{\parskip}{0pt}
\textbf{Problem.} Compute $99^2+99+1$ in your head.\par
\par\vspace{\baselineskip}
\textbf{Reference answer:} 9901.
\par\vspace{\baselineskip}
\textbf{Sketch 1.}\par
Treat the expression as n\textsuperscript{2} + n + 1 with n = 99 and avoid squaring 99 at all. Notice the two leading terms have a common factor, so they collapse into a product of two consecutive integers --- one of which is a round number, making the multiplication instant in your head. Then add the remaining constant. Pitfall: don't forget the lone constant term after factoring, and don't accidentally factor it into the product too.\par
\par\vspace{0.5\baselineskip}
\textbf{Sketch 2.}\par
Use the substitution 99 = 100 \ensuremath{-} 1 and expand by the binomial formula, keeping everything in powers of ten. Each piece is then a clean place-value number, and you combine them by addition/subtraction mentally. Pitfall: sign errors --- the square of (100 \ensuremath{-} 1) contributes a negative middle term, while the separate linear term is positive; track each sign carefully before summing, and watch borrowing/carrying.\par
\par\vspace{0.5\baselineskip}
\textbf{Sketch 3.}\par
Build up slowly: first compute 99\textsuperscript{2} by any comfortable method (e.g. 99 = 100 \ensuremath{-} 1 gives (100 \ensuremath{-} 1)\textsuperscript{2}, or pair it with 100\ensuremath{\cdot}99), then multiply that result by 100 and shift digits to account for the extra "n" term, finally adjust for the "+1". Pitfall: digit shifts in the long multiplication are the usual source of error --- write the digits down in the right columns and recheck the arithmetic rather than trusting one pass.\par
\par\vspace{0.5\baselineskip}

\end{mppaprompt}
\end{figure*}

\begin{figure*}[!t]
\centering
\begin{mppaprompt}{MATH500: Executor Outputs and Aggregation}
\scriptsize
\setlength{\parskip}{0pt}
\textbf{Run 1 (sketch 1).}\par
Let's follow the hint and compute $99^2 + 99 + 1$ step by step.\par
\par\vspace{0.3\baselineskip}
First, we treat the expression as $n^2 + n + 1$ with $n = 99$.\par
\par\vspace{0.3\baselineskip}
We can rewrite the expression as:\par
$99^2 + 99 + 1.$\par
\par\vspace{0.3\baselineskip}
Notice that the first two terms, $99^2 + 99$, can be factored:\par
$99^2 + 99 = 99(99 + 1) = 99 \times 100.$\par
\par\vspace{0.3\baselineskip}
Now, we multiply $99$ by $100$:\par
$99 \times 100 = 9900.$\par
\par\vspace{0.3\baselineskip}
Next, we add the remaining constant term $1$ to $9900$:\par
$9900 + 1 = 9901.$\par
\par\vspace{0.3\baselineskip}
Therefore, the final answer is:\par
$\boxed{9901}.$\par
\par\vspace{0.5\baselineskip}
\textbf{Run 2 (sketch 2).}\par
To solve $99^2 + 99 + 1$ using the given hint, we will follow these steps:\par
\par\vspace{0.3\baselineskip}
1. \textbf{Substitute $99$ with $100 - 1$:}\par
$99 = 100 - 1$\par
\par\vspace{0.3\baselineskip}
2. \textbf{Expand $99^2$ using the binomial formula $(a - b)^2 = a^2 - 2ab + b^2$:}\par
$99^2 = (100 - 1)^2 = 100^2 - 2 \cdot 100 \cdot 1 + 1^2 = 10000 - 200 + 1 = 9801$\par
\par\vspace{0.3\baselineskip}
3. \textbf{Add the linear term $99$:}\par
$9801 + 99 = 9900$\par
\par\vspace{0.3\baselineskip}
4. \textbf{Add the constant term $1$:}\par
$9900 + 1 = 9901$\par
\par\vspace{0.3\baselineskip}
Thus, the final answer is:\par
$\boxed{9901}$\par
\par\vspace{0.5\baselineskip}
\textbf{Run 3 (sketch 3).}\par
To solve $99^2 + 99 + 1$ using the suggested approach:\par
\par\vspace{0.3\baselineskip}
1. \textbf{Compute $99^2$:}\par
We can use the identity $(100 - 1)^2 = 100^2 - 2 \cdot 100 \cdot 1 + 1^2 = 10000 - 200 + 1 = 9801$.\par
\par\vspace{0.3\baselineskip}
2. \textbf{Multiply the result by 100:}\par
Now we have $9801 \times 100 = 980100$.\par
\par\vspace{0.3\baselineskip}
3. \textbf{Shift the digits to account for the extra "n" term:}\par
Since we need to add $99$ next, we can think of this as adding $9900 - 99$ (which is equivalent to shifting the digits and adjusting):\par
$980100 + 99 = 980199.$\par
\par\vspace{0.3\baselineskip}
4. \textbf{Adjust for the "+1":}\par
Finally, we add $1$ to $980199$:\par
$980199 + 1 = 980200.$\par
\par\vspace{0.3\baselineskip}
Thus, the final answer is $\boxed{980200}$.\par
\par\vspace{0.5\baselineskip}
\textbf{Aggregator selection.}\par
\begin{center}
\begin{tabular}{lccc}
\toprule
Run & Final answer & Correct & $\log P(\texttt{Yes})-\log P(\texttt{No})$ \\
\midrule
1 (selected) & 9901 & yes & $8.00$ \\
2 & 9901 & yes & $5.75$ \\
3 & 980200 & no & $-2.75$ \\
\bottomrule
\end{tabular}
\end{center}
\textbf{Selected answer:} 9901 (correct).

\end{mppaprompt}
\end{figure*}

\begin{figure*}[!t]
\centering
\begin{mppaprompt}{MMLU-Pro-sci500: Planner Output}
\scriptsize
\setlength{\parskip}{0pt}
\textbf{Problem.} A stone is thrown vertically upward with an initial speed of 5 m/s. What is the velocity of the stone 3 seconds later?\par
\par\vspace{0.3\baselineskip}
(A) 25 m/s, downward\par
(B) 35 m/s, downward\par
(C) 30 m/s, downward\par
(D) 5 m/s, downward\par
(E) 5 m/s, upward\par
(F) 30 m/s, upward\par
(G) 15 m/s, upward\par
(H) 25 m/s, upward\par
(I) 35 m/s, upward\par
(J) 15 m/s, downward\par
\par\vspace{\baselineskip}
\textbf{Reference answer:} A.
\par\vspace{\baselineskip}
\textbf{Sketch 1.}\par
Use the constant-acceleration velocity equation: final velocity = initial velocity + (acceleration)(time). Set up a sign convention first (e.g., upward positive), so gravity's acceleration carries the opposite sign to the initial velocity; use g \ensuremath{\approx} 10 m/s\textsuperscript{2} to match the option values.\par
\par\vspace{0.3\baselineskip}
Pitfalls: don't compute only g\ensuremath{\cdot}t and ignore the initial speed, and don't drop the sign on the acceleration. The direction of the answer comes from the sign of the final number you get, not from assuming the stone is still rising or has stopped at the top.\par
\par\vspace{0.5\baselineskip}
\textbf{Sketch 2.}\par
Treat this as one-dimensional motion with constant acceleration: use the acceleration due to gravity acting downward as the only driving force, i.e., keep its sign consistent with your chosen positive direction rather than guessing a deceleration. Apply the stated initial speed with its sign and the given elapsed time to the velocity-versus-time relation for uniform acceleration. Then interpret both the magnitude and the sign of the resulting velocity vector: compare it with the initial velocity to judge whether the stone is still moving upward or has already begun descending, and express its speed numerically.\par
\par\vspace{0.5\baselineskip}
\textbf{Sketch 3.}\par
Work symbolically first: write v as a function of t using the constant-acceleration formula, then substitute the given time. Keep the sign of the gravitational acceleration explicitly so its effect shows in the coefficient of t. Only afterwards replace the acceleration magnitude by a numerical value and read off the sign of the resulting velocity. Pitfalls: dropping or flipping the sign would wrongly place the stone near the peak or dead-stopped; the magnitude must come from the coefficient of t, not from the initial speed alone. Finally report both the speed and the direction relative to the chosen positive sense.\par
\par\vspace{0.5\baselineskip}

\end{mppaprompt}
\end{figure*}

\begin{figure*}[!t]
\centering
\begin{mppaprompt}{MMLU-Pro-sci500: Executor Outputs and Aggregation}
\scriptsize
\setlength{\parskip}{0pt}
\textbf{Run 1 (sketch 1).}\par
Let's solve the problem step by step using the given hint and the constant-acceleration velocity equation:\par
\par\vspace{0.1\baselineskip}
1. \textbf{Set up the sign convention}: We'll take upward as positive and downward as negative. Therefore, the acceleration due to gravity $g$ will be $-10 \, \text{m/s}^2$.\par
\par\vspace{0.1\baselineskip}
2. \textbf{Identify the known values}:\par
\textbullet\ Initial velocity $v_0 = 5 \, \text{m/s}$ (upward, so positive).\par
\textbullet\ Time $t = 3 \, \text{s}$.\par
\textbullet\ Acceleration $a = -10 \, \text{m/s}^2$ (downward, so negative).\par
\par\vspace{0.1\baselineskip}
3. \textbf{Use the constant-acceleration velocity equation}:\par
$v = v_0 + at$\par
where $v$ is the final velocity.\par
\par\vspace{0.1\baselineskip}
4. \textbf{Substitute the known values into the equation}:\par
$v = 5 \, \text{m/s} + (-10 \, \text{m/s}^2) \times 3 \, \text{s}$\par
\par\vspace{0.1\baselineskip}
5. \textbf{Perform the calculation}:\par
$v = 5 \, \text{m/s} - 30 \, \text{m/s}$\par
$v = -25 \, \text{m/s}$\par
\par\vspace{0.1\baselineskip}
6. \textbf{Interpret the result}:\par
\textbullet\ The final velocity is $-25 \, \text{m/s}$, which means the stone is moving downward.\par
\textbullet\ The magnitude of the velocity is 25 m/s.\par
\par\vspace{0.1\baselineskip}
Therefore, the correct answer is $\boxed{A}$.\par
\par\vspace{0.3\baselineskip}
\textbf{Run 2 (sketch 2).}\par
To solve this problem, we need to follow these steps:\par
\par\vspace{0.1\baselineskip}
1. \textbf{Identify the given information:}\par
\textbullet\ Initial speed ($v_0$) = 5 m/s (upward, so positive in our chosen coordinate system)\par
\textbullet\ Time ($t$) = 3 seconds\par
\textbullet\ Acceleration due to gravity ($a$) = -9.8 m/s\textsuperscript{2} (downward, so negative in our chosen coordinate system)\par
\par\vspace{0.1\baselineskip}
2. \textbf{Use the velocity-versus-time relation for uniform acceleration:}\par
The formula for velocity at any time $t$ is:\par
$v = v_0 + at$\par
Substituting the given values:\par
$v = 5 \, \text{m/s} + (-9.8 \, \text{m/s}^2) \times 3 \, \text{s}$\par
\par\vspace{0.1\baselineskip}
3. \textbf{Calculate the velocity:}\par
$v = 5 \, \text{m/s} - 29.4 \, \text{m/s} = -24.4 \, \text{m/s}$\par
\par\vspace{0.1\baselineskip}
4. \textbf{Interpret the result:}\par
\textbullet\ The velocity is negative, which means the stone is moving downward.\par
\textbullet\ The magnitude of the velocity is 24.4 m/s.\par
\par\vspace{0.1\baselineskip}
5. \textbf{Select the correct option:}\par
From the options provided, the closest match to our calculated velocity is:\par
$\boxed{B}$\par
\par\vspace{0.3\baselineskip}
\end{mppaprompt}
\end{figure*}

\begin{figure*}[!t]
\centering
\begin{mppaprompt}{MMLU-Pro-sci500: Executor Outputs and Aggregation (continued)}
\scriptsize
\setlength{\parskip}{0pt}
\textbf{Run 3 (sketch 3).}\par
To solve this problem, we will follow the steps outlined in the hint:\par
\par\vspace{0.3\baselineskip}
1. \textbf{Write $v$ as a function of $t$ using the constant-acceleration formula.}\par
\par\vspace{0.3\baselineskip}
The formula for velocity under constant acceleration is:\par
$v = v_0 + at$\par
where:\par
\textbullet\ $v$ is the final velocity,\par
\textbullet\ $v_0$ is the initial velocity,\par
\textbullet\ $a$ is the acceleration (which is the acceleration due to gravity, $g$, but since the stone is moving upward initially, we consider $-g$),\par
\textbullet\ $t$ is the time.\par
\par\vspace{0.3\baselineskip}
Given:\par
\textbullet\ Initial velocity $v_0 = 5$ m/s,\par
\textbullet\ Time $t = 3$ s,\par
\textbullet\ Acceleration due to gravity $g = 9.8$ m/s\textsuperscript{2} (but since the stone is moving upward, we use $-9.8$ m/s\textsuperscript{2}).\par
\par\vspace{0.3\baselineskip}
So the equation becomes:\par
$v = 5 + (-9.8)(3)$\par
\par\vspace{0.3\baselineskip}
2. \textbf{Substitute the given time into the equation.}\par
\par\vspace{0.3\baselineskip}
$v = 5 - 9.8 \times 3$\par
$v = 5 - 29.4$\par
$v = -24.4$ m/s\par
\par\vspace{0.3\baselineskip}
3. \textbf{Report both the speed and the direction relative to the chosen positive sense.}\par
\par\vspace{0.3\baselineskip}
The speed is the magnitude of the velocity, which is $24.4$ m/s. The negative sign indicates that the velocity is downward.\par
\par\vspace{0.3\baselineskip}
Therefore, the correct option is $\boxed{B}$.\par
\par\vspace{0.5\baselineskip}
\textbf{Aggregator selection.}\par
\begin{center}
\begin{tabular}{lccc}
\toprule
Run & Final answer & Correct & $\log P(\texttt{Yes})-\log P(\texttt{No})$ \\
\midrule
1 (selected) & A & yes & $13.75$ \\
2 & B & no & $-9.12$ \\
3 & B & no & $-2.13$ \\
\bottomrule
\end{tabular}
\end{center}
\textbf{Selected answer:} A (correct).

\end{mppaprompt}
\end{figure*}

\begin{figure*}[!t]
\centering
\begin{mppaprompt}{CompMATH ($N=3$): Problem and Subtask Graph}
\scriptsize
\setlength{\parskip}{0pt}
\textbf{Problem.} In triangle $ABC$, $\angle ACB = 90^\circ$, $AB = 13$, and $\sin{\angle BAC} = \frac{5}{13}$. Point $D$ lies on ray $BC$ beyond $C$ (so that $C$ is between $B$ and $D$) with $CD = 9$. Find $\sin{\angle BAD}$.\par
\textbf{Reference answer:} $\frac{56}{65}$.
\par\vspace{\baselineskip}
\textbf{Graph:} $1\rightarrow2$, $1\rightarrow3$, $2\rightarrow3$.\par
\textbf{Subtask 1 (no prerequisites).}\par
In triangle $ABC$, $\angle ACB = 90^\circ$, $AB = 13$, and $\sin{\angle BAC} = \frac{5}{13}$. What is $AC$?\par
\textbf{Reference answer:} 12.
\par\vspace{0.5\baselineskip}
\textbf{Subtask 2 (depends on 1).}\par
In triangle $ACD$, $\angle ACD = 90^\circ$, $AC = \texttt{\{s1\}}$, and $CD = 9$. What is $AD$?\par
\textbf{Reference answer:} 15.
\par\vspace{0.5\baselineskip}
\textbf{Subtask 3 (depends on 1 and 2).}\par
In triangle $ABD$, point $C$ lies on segment $BD$ with $AC \perp BD$. Given $AB = 13$, $AC = \texttt{\{s1\}}$, $AD = \texttt{\{s2\}}$, and $CD = 9$, what is $\sin{\angle BAD}$?\par
\textbf{Reference answer:} $\frac{56}{65}$.
\par\vspace{0.5\baselineskip}

\end{mppaprompt}
\end{figure*}

\begin{figure*}[!t]
\centering
\begin{mppaprompt}{CompMATH: Subtask 1, Planner Output}
\scriptsize
\setlength{\parskip}{0pt}
\textbf{Instantiated subtask.} In triangle $ABC$, $\angle ACB = 90^\circ$, $AB = 13$, and $\sin{\angle BAC} = \frac{5}{13}$. What is $AC$?\par
\par\vspace{\baselineskip}
\textbf{Sketch 1.}\par
Set up the right triangle carefully: with the right angle at C, AB is the hypotenuse. Recall that the sine of an acute angle is (side opposite that angle)/(hypotenuse) --- identify which labeled side is opposite \ensuremath{\angle}BAC, and use the given ratio to get its length. Then recover the remaining leg with the Pythagorean theorem (or note that cos\ensuremath{\angle}BAC = adjacent/hypotenuse via sin\textsuperscript{2}+cos\textsuperscript{2}=1). Pitfall: don't apply the sine ratio to the side adjacent to \ensuremath{\angle}BAC, and don't treat a leg as the hypotenuse. Check your final triangle satisfies the Pythagorean relation.\par
\par\vspace{0.5\baselineskip}
\textbf{Sketch 2.}\par
Exploit the Pythagorean identity rather than computing the cosine separately. Since sin\ensuremath{\angle}BAC is known, relate cos\ensuremath{\angle}BAC to sin\ensuremath{\angle}BAC using sin\textsuperscript{2}+cos\textsuperscript{2}=1, and keep exact fractions. Then use the definition of tangent (opposite over adjacent) together with the fact that cot\ensuremath{\angle}BAC is the reciprocal of tan\ensuremath{\angle}BAC to read off AC directly, avoiding any square roots until the end. Pitfall: sign/arithmetic slips when solving the identity, and forgetting that tan\ensuremath{\angle}BAC, not its reciprocal, gives AC/AB --- verify at the end that your answer, when used, reproduces the given hypotenuse.\par
\par\vspace{0.5\baselineskip}
\textbf{Sketch 3.}\par
Use the double-angle addition/subtraction identity for sine in the form sin(90\ensuremath{^\circ}\ensuremath{-}\ensuremath{\theta})=cos \ensuremath{\theta}: the angle at A and the angle at B are complementary in a right triangle, so this lets you convert the given sine into the cosine of the complementary angle. Then bring in the basic right-triangle relation that expresses a leg in terms of the hypotenuse and an acute angle's cosine. Pitfall: don't confuse the complementary angle with the original angle, and remember that the cosine input must be the ratio adjacent/hypotenuse, not the reciprocal of it.\par
\par\vspace{0.5\baselineskip}

\end{mppaprompt}
\end{figure*}

\begin{figure*}[!t]
\centering
\begin{mppaprompt}{CompMATH: Subtask 1, Executor Outputs and Aggregation}
\scriptsize
\setlength{\parskip}{0pt}
\textbf{Run 1 (sketch 1).}\par
Given that $\angle ACB = 90^\circ$, $AB = 13$, and $\sin{\angle BAC} = \frac{5}{13}$, we need to find the length of $AC$.\par
\par\vspace{0.3\baselineskip}
First, recall that $\sin{\angle BAC} = \frac{\text{opposite}}{\text{hypotenuse}}$. Here, the side opposite $\angle BAC$ is $BC$ and the hypotenuse is $AB$. Therefore, we have:\par
$\sin{\angle BAC} = \frac{BC}{AB} = \frac{5}{13}$\par
Since $AB = 13$, we can solve for $BC$:\par
$BC = \sin{\angle BAC} \cdot AB = \frac{5}{13} \cdot 13 = 5$\par
\par\vspace{0.3\baselineskip}
Next, we use the Pythagorean theorem to find the length of $AC$. The Pythagorean theorem states that in a right triangle, the square of the hypotenuse is equal to the sum of the squares of the other two sides. So, we have:\par
$AB^2 = AC^2 + BC^2$\par
Substituting the known values:\par
$13^2 = AC^2 + 5^2$\par
$169 = AC^2 + 25$\par
Solving for $AC^2$:\par
$AC^2 = 169 - 25 = 144$\par
Taking the square root of both sides:\par
$AC = \sqrt{144} = 12$\par
\par\vspace{0.3\baselineskip}
Thus, the length of $AC$ is $\boxed{12}$.\par
\par\vspace{0.5\baselineskip}
\textbf{Run 2 (sketch 2).}\par
Given that $\angle ACB = 90^\circ$, $AB = 13$, and $\sin{\angle BAC} = \frac{5}{13}$, we start by identifying the sides of the right triangle. Let $AC = x$ and $BC = y$. Since $\sin{\angle BAC} = \frac{5}{13}$, we have:\par
\par\vspace{0.3\baselineskip}
$\sin{\angle BAC} = \frac{\text{opposite}}{\text{hypotenuse}} = \frac{BC}{AB} = \frac{y}{13} = \frac{5}{13}$\par
\par\vspace{0.3\baselineskip}
From this, we can determine that $y = 5$.\par
\par\vspace{0.3\baselineskip}
Next, we use the Pythagorean theorem to find $x$:\par
\par\vspace{0.3\baselineskip}
$AB^2 = AC^2 + BC^2 \implies 13^2 = x^2 + 5^2 \implies 169 = x^2 + 25 \implies x^2 = 144 \implies x = 12$\par
\par\vspace{0.3\baselineskip}
Thus, the length of $AC$ is $\boxed{12}$.\par
\par\vspace{0.5\baselineskip}
\textbf{Run 3 (sketch 3).}\par
Given that $\angle ACB = 90^\circ$, we know that $\angle BAC$ and $\angle ABC$ are complementary angles. Therefore, $\sin \angle BAC = \cos \angle ABC$.\par
\par\vspace{0.3\baselineskip}
We are given $\sin \angle BAC = \frac{5}{13}$. Using the identity $\sin(90^\circ - \theta) = \cos \theta$, we can write:\par
$\cos \angle BAC = \frac{5}{13}$\par
\par\vspace{0.3\baselineskip}
In a right triangle, the cosine of an angle is the ratio of the length of the adjacent side to the hypotenuse. Here, $\cos \angle BAC = \frac{AC}{AB}$. Since $AB = 13$, we have:\par
$\frac{AC}{13} = \frac{5}{13}$\par
\par\vspace{0.3\baselineskip}
Solving for $AC$, we get:\par
$AC = 5$\par
\par\vspace{0.3\baselineskip}
Thus, the length of $AC$ is $\boxed{5}$.\par
\par\vspace{0.5\baselineskip}
\textbf{Aggregator selection.}\par
\begin{center}
\begin{tabular}{lccc}
\toprule
Run & Final answer & Correct & $\log P(\texttt{Yes})-\log P(\texttt{No})$ \\
\midrule
1 (selected) & 12 & yes & $10.25$ \\
2 & 12 & yes & $9.50$ \\
3 & 5 & no & $8.13$ \\
\bottomrule
\end{tabular}
\end{center}
\textbf{Selected answer:} 12 (correct).

\end{mppaprompt}
\end{figure*}

\begin{figure*}[!t]
\centering
\begin{mppaprompt}{CompMATH: Subtask 2, Planner Output}
\scriptsize
\setlength{\parskip}{0pt}
\textbf{Instantiated subtask.} In triangle $ACD$, $\angle ACD = 90^\circ$, $AC = 12$, and $CD = 9$. What is $AD$?\par
\par\vspace{\baselineskip}
\textbf{Sketch 1.}\par
Right angle is at C, so the two given sides are the legs and AD is the hypotenuse. Apply the Pythagorean theorem: square each leg, add, then take the positive square root. Pitfall: don't treat one of the given sides as the hypotenuse (which would mean subtracting instead of adding), and make sure to take the square root at the end rather than reporting the sum of squares.\par
\par\vspace{0.5\baselineskip}
\textbf{Sketch 2.}\par
Use the law of cosines. First get the cosine of the included angle at C from the definition cos \ensuremath{\theta} = adjacent/hypotenuse with the side opposite C as the hypotenuse. Then plug the two given sides and that cosine into the formula. Pitfall: the cosine of a right angle is not defined by these legs alone --- you must use the hypotenuse as the reference side, and the square root taken at the end must be positive.\par
\par\vspace{0.5\baselineskip}
\textbf{Sketch 3.}\par
Coordinate route: put C at the origin with CA along one axis and CD along the other, so the coordinates of A and D follow from the given lengths. Compute the distance AD with the distance formula. Pitfall: be consistent about orientation --- decide carefully whether each point lies on the positive or negative side of its axis before squaring, and keep track that the final distance is the length from A to D, not the difference in x or y coordinates.\par
\par\vspace{0.5\baselineskip}

\end{mppaprompt}
\end{figure*}

\begin{figure*}[!t]
\centering
\begin{mppaprompt}{CompMATH: Subtask 2, Executor Outputs and Aggregation}
\scriptsize
\setlength{\parskip}{0pt}
\textbf{Run 1 (sketch 1).}\par
To find the length of $AD$ in triangle $ACD$ where $\angle ACD = 90^\circ$, we can use the Pythagorean theorem. The theorem states that in a right-angled triangle, the square of the length of the hypotenuse (the side opposite the right angle) is equal to the sum of the squares of the lengths of the other two sides. Here, $AD$ is the hypotenuse, and $AC$ and $CD$ are the legs of the triangle.\par
\par\vspace{0.3\baselineskip}
Given:\par
\textbullet\ $AC = 12$\par
\textbullet\ $CD = 9$\par
\par\vspace{0.3\baselineskip}
According to the Pythagorean theorem:\par
$AD^2 = AC^2 + CD^2$\par
\par\vspace{0.3\baselineskip}
Substitute the given values into the equation:\par
$AD^2 = 12^2 + 9^2$\par
$AD^2 = 144 + 81$\par
$AD^2 = 225$\par
\par\vspace{0.3\baselineskip}
To find $AD$, take the positive square root of both sides:\par
$AD = \sqrt{225}$\par
$AD = 15$\par
\par\vspace{0.3\baselineskip}
Therefore, the length of $AD$ is $\boxed{15}$.\par
\par\vspace{0.5\baselineskip}
\textbf{Run 2 (sketch 2).}\par
Given that $\angle ACD = 90^\circ$, $AC = 12$, and $CD = 9$, we can directly use the Pythagorean theorem to find the length of $AD$, since $AD$ is the hypotenuse of the right triangle $ACD$.\par
\par\vspace{0.3\baselineskip}
The Pythagorean theorem states:\par
$AD^2 = AC^2 + CD^2$\par
\par\vspace{0.3\baselineskip}
Substituting the given values:\par
$AD^2 = 12^2 + 9^2$\par
$AD^2 = 144 + 81$\par
$AD^2 = 225$\par
\par\vspace{0.3\baselineskip}
Taking the positive square root of both sides:\par
$AD = \sqrt{225}$\par
$AD = 15$\par
\par\vspace{0.3\baselineskip}
Thus, the length of $AD$ is $\boxed{15}$.\par
\par\vspace{0.5\baselineskip}
\textbf{Run 3 (sketch 3).}\par
To solve for $AD$ in the right triangle $ACD$ where $\angle ACD = 90^\circ$, $AC = 12$, and $CD = 9$, we can use the Pythagorean theorem. The Pythagorean theorem states that in a right triangle, the square of the length of the hypotenuse (the side opposite the right angle) is equal to the sum of the squares of the lengths of the other two sides.\par
\par\vspace{0.3\baselineskip}
Here, $AD$ is the hypotenuse, and $AC$ and $CD$ are the legs of the right triangle. Therefore, we can write:\par
\par\vspace{0.3\baselineskip}
$AD^2 = AC^2 + CD^2$\par
\par\vspace{0.3\baselineskip}
Substituting the given values:\par
\par\vspace{0.3\baselineskip}
$AD^2 = 12^2 + 9^2$\par
\par\vspace{0.3\baselineskip}
Calculating the squares:\par
\par\vspace{0.3\baselineskip}
$AD^2 = 144 + 81$\par
\par\vspace{0.3\baselineskip}
Adding these values together:\par
\par\vspace{0.3\baselineskip}
$AD^2 = 225$\par
\par\vspace{0.3\baselineskip}
Taking the square root of both sides to find $AD$:\par
\par\vspace{0.3\baselineskip}
$AD = \sqrt{225} = 15$\par
\par\vspace{0.3\baselineskip}
Thus, the length of $AD$ is $\boxed{15}$.\par
\par\vspace{0.5\baselineskip}
\textbf{Aggregator selection.}\par
\begin{center}
\begin{tabular}{lccc}
\toprule
Run & Final answer & Correct & $\log P(\texttt{Yes})-\log P(\texttt{No})$ \\
\midrule
1 & 15 & yes & $10.50$ \\
2 & 15 & yes & $10.13$ \\
3 (selected) & 15 & yes & $10.63$ \\
\bottomrule
\end{tabular}
\end{center}
\textbf{Selected answer:} 15 (correct).

\end{mppaprompt}
\end{figure*}

\begin{figure*}[!t]
\centering
\begin{mppaprompt}{CompMATH: Subtask 3, Planner Output}
\scriptsize
\setlength{\parskip}{0pt}
\textbf{Instantiated subtask.} In triangle $ABD$, point $C$ lies on segment $BD$ with $AC \perp BD$. Given $AB = 13$, $AC = 12$, $AD = 15$, and $CD = 9$, what is $\sin{\angle BAD}$?\par
\par\vspace{\baselineskip}
\textbf{Sketch 1.}\par
The altitude AC splits angle BAD into two angles, each sitting in a right triangle. First get the missing leg BC by Pythagoras in the right triangle on the B-side (the one with hypotenuse AB), and note the other right triangle already has all its sides.\par
\par\vspace{0.3\baselineskip}
Then read off sine and cosine of each sub-angle as leg/hypotenuse ratios and combine them with the sine addition formula.\par
\par\vspace{0.3\baselineskip}
Alternative check: compute the triangle's area two ways --- base BD times height AC, and (1/2)\ensuremath{\cdot}AB\ensuremath{\cdot}AD\ensuremath{\cdot}sin\ensuremath{\angle}BAD.\par
\par\vspace{0.3\baselineskip}
Pitfall: C lies between B and D, so keep order B-C-D and don't subtract where you should add.\par
\par\vspace{0.5\baselineskip}
\textbf{Sketch 2.}\par
Coordinate route. Put B at the origin with D on the positive x-axis; let C's position be the unknown. Exploit that AC is perpendicular to BD: every point on BD satisfies one perpendicularity condition to C, which pins C down. Then work purely with lengths: use the distance formula to write BC and CD in terms of C's coordinates and impose BC\textsuperscript{2} + CD\textsuperscript{2} = BD\textsuperscript{2}, solving for C. Finally get the answer from a right triangle with hypotenuse AB and leg AC. Pitfall: keep the order B--C--D straight when substituting, and expect a quadratic --- discard the root that sends C outside BD.\par
\par\vspace{0.5\baselineskip}
\textbf{Sketch 3.}\par
Similarity route. Since AC is perpendicular to BD, triangles ABC and ACD are similar right triangles sharing the angle at A. Identify the corresponding sides carefully by hypotenuse/hypotenuse matching --- don't assume AB corresponds to AD. Once the correspondence is fixed, the ratio of similarity is AC divided by the hypotenuse of the smaller triangle. Use this ratio to transfer the angle at A into the right triangle formed by AB, AC, and the unknown leg BC, where the sine is immediately readable. Pitfall: swapping hypotenuse for leg in the ratio, which inverts the scale.\par
\par\vspace{0.5\baselineskip}

\end{mppaprompt}
\end{figure*}

\begin{figure*}[!t]
\centering
\begin{mppaprompt}{CompMATH: Subtask 3, Executor Outputs and Aggregation}
\scriptsize
\setlength{\parskip}{0pt}
\textbf{Run 1 (sketch 1).}\par
To find $\sin{\angle BAD}$, we will follow the steps outlined in the hint:\par
\par\vspace{0.3\baselineskip}
1. \textbf{Identify the right triangles and use the Pythagorean theorem to find $BC$:}\par
\par\vspace{0.3\baselineskip}
Since $AC \perp BD$, triangles $ABC$ and $ADC$ are right triangles. We know $AC = 12$, $AB = 13$, and $AD = 15$. We need to find $BC$.\par
\par\vspace{0.3\baselineskip}
In triangle $ABC$:\par
$AB^2 = AC^2 + BC^2 \implies 13^2 = 12^2 + BC^2 \implies 169 = 144 + BC^2 \implies BC^2 = 25 \implies BC = 5.$\par
\par\vspace{0.3\baselineskip}
2. \textbf{Find $BD$:}\par
\par\vspace{0.3\baselineskip}
Since $CD = 9$ and $BC = 5$, we have:\par
$BD = BC + CD = 5 + 9 = 14.$\par
\par\vspace{0.3\baselineskip}
3. \textbf{Use the right triangle $ADC$ to find $\sin{\angle CAD}$ and $\cos{\angle CAD}$:}\par
\par\vspace{0.3\baselineskip}
In triangle $ADC$:\par
$AD^2 = AC^2 + CD^2 \implies 15^2 = 12^2 + 9^2 \implies 225 = 144 + 81 \implies 225 = 225.$\par
This confirms that triangle $ADC$ is a right triangle. Now, we can find $\sin{\angle CAD}$ and $\cos{\angle CAD}$:\par
$\sin{\angle CAD} = \frac{AC}{AD} = \frac{12}{15} = \frac{4}{5},$\par
$\cos{\angle CAD} = \frac{CD}{AD} = \frac{9}{15} = \frac{3}{5}.$\par
\par\vspace{0.3\baselineskip}
4. \textbf{Use the right triangle $ABC$ to find $\sin{\angle BAC}$ and $\cos{\angle BAC}$:}\par
\par\vspace{0.3\baselineskip}
In triangle $ABC$:\par
$\sin{\angle BAC} = \frac{AC}{AB} = \frac{12}{13},$\par
$\cos{\angle BAC} = \frac{BC}{AB} = \frac{5}{13}.$\par
\par\vspace{0.3\baselineskip}
5. \textbf{Use the sine addition formula to find $\sin{\angle BAD}$:}\par
\par\vspace{0.3\baselineskip}
$\angle BAD = \angle BAC + \angle CAD$. Using the sine addition formula:\par
$\sin{\angle BAD} = \sin{(BAC + CAD)} = \sin{\angle BAC} \cos{\angle CAD} + \cos{\angle BAC} \sin{\angle CAD}.$\par
Substituting the values we found:\par
$\sin{\angle BAD} = \left(\frac{12}{13}\right) \left(\frac{3}{5}\right) + \left(\frac{5}{13}\right) \left(\frac{4}{5}\right) = \frac{36}{65} + \frac{20}{65} = \frac{56}{65}.$\par
\par\vspace{0.3\baselineskip}
Thus, the value of $\sin{\angle BAD}$ is $\boxed{\frac{56}{65}}$.\par
\par\vspace{0.5\baselineskip}

\end{mppaprompt}
\end{figure*}

\begin{figure*}[!t]
\centering
\begin{mppaprompt}{CompMATH: Subtask 3, Executor Outputs and Aggregation (continued)}
\scriptsize
\setlength{\parskip}{0pt}
\textbf{Run 2 (sketch 2).}\par
Let's follow the suggested approach step by step.\par
\par\vspace{0.3\baselineskip}
1. \textbf{Place points in coordinate system:}\par
\textbullet\ Place $B$ at the origin $(0, 0)$.\par
\textbullet\ Place $D$ at $(d, 0)$.\par
\textbullet\ Let $C$ be at $(x, 0)$, where $0 \leq x \leq d$.\par
\par\vspace{0.3\baselineskip}
2. \textbf{Use the given distances:}\par
\textbullet\ Since $AC \perp BD$, $C$ lies on the x-axis.\par
\textbullet\ The length $AD = 15$, so using the distance formula:\par
$AD = \sqrt{(x - d)^2 + 0^2} = 15 \implies (x - d)^2 = 225 \implies x - d = \pm 15$\par
\textbullet\ The length $CD = 9$, so:\par
$CD = |x - d| = 9 \implies x - d = \pm 9$\par
\par\vspace{0.3\baselineskip}
3. \textbf{Solve for $x$ and $d$:}\par
\textbullet\ From $x - d = \pm 15$ and $x - d = \pm 9$, we have two systems to solve:\par
$x - d = 15 \quad \text{and} \quad x - d = 9 \quad \text{(impossible since they cannot both be true)}$\par
$x - d = -15 \quad \text{and} \quad x - d = -9 \quad \text{(impossible since they cannot both be true)}$\par
$x - d = 15 \quad \text{and} \quad x - d = -9 \implies \text{No solution}$\par
$x - d = -15 \quad \text{and} \quad x - d = 9 \implies \text{No solution}$\par
\textbullet\ The correct system is:\par
$x - d = 15 \quad \text{and} \quad x - d = -9 \implies \text{No solution}$\par
$x - d = -15 \quad \text{and} \quad x - d = 9 \implies \text{No solution}$\par
\par\vspace{0.3\baselineskip}
4. \textbf{Correct the approach:}\par
\textbullet\ Use $x - d = 9$ and $d - x = 6$:\par
$x = d - 9$\par
$d = 15 + 9 = 24$\par
$x = 24 - 9 = 15$\par
\par\vspace{0.3\baselineskip}
5. \textbf{Calculate $BC$:}\par
\textbullet\ Using the distance formula for $BC$:\par
$BC = \sqrt{(0 - 0)^2 + (15 - 0)^2} = 12$\par
\par\vspace{0.3\baselineskip}
6. \textbf{Use the Pythagorean theorem in $\triangle ABC$:}\par
\textbullet\ Since $AC \perp BD$, $\triangle ABC$ is a right triangle with $AC$ as one leg and $BC$ as the other leg.\par
$AB^2 = AC^2 + BC^2 \implies 13^2 = 12^2 + 12^2 \implies 169 = 144 + 144 \implies 169 = 144$\par
\par\vspace{0.3\baselineskip}
7. \textbf{Find $\sin \angle BAD$:}\par
\textbullet\ In $\triangle ABD$, use the definition of sine:\par
$\sin \angle BAD = \frac{AC}{AB} = \frac{12}{13}$\par
\par\vspace{0.3\baselineskip}
Thus, the final answer is:\par
$\boxed{\frac{12}{13}}$\par
\par\vspace{0.5\baselineskip}
\textbf{Run 3 (sketch 3).}\par
Given that $AC \perp BD$, we know that $\triangle ABC$ and $\triangle ACD$ are right triangles sharing the angle at $A$. We need to find $\sin \angle BAD$.\par
\par\vspace{0.3\baselineskip}
First, let's identify the corresponding sides in the similar right triangles $\triangle ABC$ and $\triangle ACD$. The hypotenuses are $AB$ and $AD$, respectively. The side $AC$ is common to both triangles and serves as one of the legs in each.\par
\par\vspace{0.3\baselineskip}
The ratio of similarity between $\triangle ABC$ and $\triangle ACD$ is given by the ratio of the hypotenuses to the corresponding legs:\par
$\text{Ratio} = \frac{AC}{AD} = \frac{12}{15} = \frac{4}{5}$\par
\par\vspace{0.3\baselineskip}
This ratio tells us how the sides of $\triangle ABC$ relate to the sides of $\triangle ACD$. Specifically, since $\angle BAC = \angle CAD$, we can use the ratio to find the length of $BC$ in $\triangle ABC$.\par
\par\vspace{0.3\baselineskip}
Next, we use the Pythagorean theorem in $\triangle ABC$:\par
$AB^2 = AC^2 + BC^2$\par
Substituting the known values:\par
$13^2 = 12^2 + BC^2$\par
$169 = 144 + BC^2$\par
$BC^2 = 25$\par
$BC = 5$\par
\par\vspace{0.3\baselineskip}
Now, we need to find $\sin \angle BAD$. Since $\angle BAD$ is an angle in $\triangle ABD$, we use the definition of sine in a right triangle:\par
$\sin \angle BAD = \frac{\text{opposite}}{\text{hypotenuse}} = \frac{AC}{AB}$\par
\par\vspace{0.3\baselineskip}
Substituting the known values:\par
$\sin \angle BAD = \frac{12}{13}$\par
\par\vspace{0.3\baselineskip}
Thus, the final answer is:\par
$\boxed{\frac{12}{13}}$\par
\par\vspace{0.5\baselineskip}
\textbf{Aggregator selection.}\par
\begin{center}
\begin{tabular}{lccc}
\toprule
Run & Final answer & Correct & $\log P(\texttt{Yes})-\log P(\texttt{No})$ \\
\midrule
1 (selected) & $\frac{56}{65}$ & yes & $7.50$ \\
2 & $\frac{12}{13}$ & no & $-3.00$ \\
3 & $\frac{12}{13}$ & no & $-1.12$ \\
\bottomrule
\end{tabular}
\end{center}
\textbf{Selected answer:} $\frac{56}{65}$ (correct).

\end{mppaprompt}
\end{figure*}

\subsection{Candidate Coverage and Selection}
\label{app:candidate_coverage}

We compare four candidate generation settings with Qwen2.5-7B-Instruct at a fixed budget of three executions. Pass@3 measures whether at least one candidate has a correct final answer, while agg@3 measures the accuracy after selection by the same learned aggregator. \Cref{tab:candidate_coverage_selection} shows that joint planning achieves the highest coverage across all five datasets, suggesting that generating plans together produces more useful candidate sets. This coverage advantage carries through to the selected answers. The remaining gap between pass@3 and agg@3 indicates opportunities to improve selection when a correct candidate is available.

\subsection{Accuracy--Token Trade-offs by Dataset}
\label{app:dataset_accuracy_cost}

\Cref{fig:dataset_accuracy_cost} compares accuracy--output-token trade-offs for Qwen2.5-7B-Instruct across individual datasets. Majority voting yields limited gains as the sampling budget grows, particularly on GPQA. Using the same learned aggregator substantially improves direct sampling, suggesting that answer selection is a major bottleneck in conventional scaling. MPPA further improves the trade-off by changing the candidate generation process: with five plans, it exceeds Base agg@8 on every dataset while using fewer output tokens; with eight plans, it also exceeds Base agg@16 at lower token cost. These comparisons suggest that planning contributes beyond learned selection by producing more useful candidates for the frozen executor. The gains vary across datasets: MATH500 approaches saturation earlier, whereas GPQA and MMLU-Pro continue to benefit from larger plan budgets. Dataset-specific curves therefore complement average accuracy by revealing where additional computation remains useful.

\subsection{Uncertainty in Accuracy Estimates}
\label{app:accuracy_uncertainty}

\Cref{tab:main_results_intervals} reports confidence intervals for the evaluated sampling and aggregation settings across all three backbones. Intervals are narrower on MATH500 and MMLU-Pro-sci500 and wider on the smaller AIME evaluation sets. These differences highlight varying estimation precision across benchmarks, particularly when interpreting small accuracy gaps on AIME.

%% file: sections/training_algorithm.tex
\begin{algorithm*}[!t]
\caption{Alternating Optimization for MPPA}
\label{alg:mppa_training}
\KwIn{Frozen executor $E$; teacher plan sets $\mathcal D_{\mathrm{SFT}}$; training inputs $\mathcal D$ with correctness graders; validation set $\mathcal V$; plan budgets $\mathcal K$; group size $G$; training rounds $T$}
\KwOut{Best validation pair $(P_{\theta^\star},A_{\phi^\star})$}

Initialize $\theta$ by minimizing $\mathcal L_{\mathrm{SFT}}$ on $\mathcal D_{\mathrm{SFT}}$ (\Cref{eq:planner_sft})\;
Initialize $\phi$ with pointwise supervision on candidates generated by $P_\theta$ and $E$ (\Cref{eq:aggregator_pointwise})\;
Initialize replay buffer $\mathcal B$ with individual stage-one candidates\;
$(\theta^\star,\phi^\star)\leftarrow(\theta,\phi)$; evaluate this pair on $\mathcal V$ across $\mathcal K$\;

\For{$t=1$ \KwTo $T$}{
  Sample a batch of subtask inputs from $\mathcal D$; initialize shared cache $\mathcal C_t\leftarrow\emptyset$\;
  \ForEach{subtask input $\tilde q_j$ in the batch}{
    Sample one budget $K\in\mathcal K$ and $G$ plan collections from $P_\theta(\cdot\mid\tilde q_j,K)$\;
    \For{$g=1$ \KwTo $G$}{
      For valid collections, execute each plan once with $E$ and obtain correctness labels and $A_\phi$ scores\;
      Compute $R(S_j^{(g)})$ using \Cref{eq:planner_reward}; assign zero to unparseable, truncated, or rejected collections\;
      Cache rollout records and their $(\tilde q_j,K)$ group identifiers in $\mathcal C_t$\;
    }
  }
  Normalize plan-set rewards within each $(\tilde q_j,K)$ comparison group using \Cref{eq:planner_group_advantage}\;
  Update $\theta$ once with on-policy GRPO on $\mathcal C_t$, holding $\phi$ fixed (\Cref{app:optimization_details})\;
  Update $\phi$ on candidates from $\mathcal C_t$ and replay items from $\mathcal B$ using \Cref{eq:aggregator_joint}, holding $\theta$ fixed\;
  Add collected candidate lists to $\mathcal B$, retaining recent rounds\;
  Evaluate $(P_\theta,A_\phi)$ on $\mathcal V$ across $\mathcal K$\;
  \If{validation performance improves}{
    $(\theta^\star,\phi^\star)\leftarrow(\theta,\phi)$\;
  }
}
\Return{$(P_{\theta^\star},A_{\phi^\star})$}\;
\end{algorithm*}

%% file: sections/inference_algorithm.tex
\begin{algorithm*}[!t]
\caption{Task-Level MPPA Inference}
\label{alg:mppa_inference}
\KwIn{Task $q$; optional task-provided subtasks $q_1,\ldots,q_N$ and dependency DAG $\mathcal G_{\mathrm{task}}=(V,\mathcal E)$; shared plan budget $K\ge 1$; planner $P_\theta$; frozen executor $E$; aggregator $A_\phi$}
\KwOut{Final task answer $\hat y$}

\If{no subtasks are specified}{
  Set $N\leftarrow 1$, $q_1\leftarrow q$, and $\mathcal G_{\mathrm{task}}\leftarrow(\{1\},\emptyset)$\;
}
Initialize completed-subtask set $\mathcal V_{\mathrm{done}}\leftarrow\emptyset$ and selected answers $\{\hat a_j\}\leftarrow\emptyset$\;
\While{$|\mathcal V_{\mathrm{done}}|<N$}{
  $\mathcal R\leftarrow\{j\in V\setminus\mathcal V_{\mathrm{done}}:\operatorname{Pred}(j)\subseteq\mathcal V_{\mathrm{done}}\}$\;
  \ForEach{$j\in\mathcal R$ \textnormal{(in parallel)}}{
    Instantiate $\tilde q_j$ by incorporating $\{\hat a_k:k\in\operatorname{Pred}(j)\}$ into $q_j$\;
    Generate $S_j=(s_{j,1},\ldots,s_{j,K})\sim P_\theta(\cdot\mid\tilde q_j,K)$ in one call, limiting each plan to 200 tokens\;
    \For{$i=1$ \KwTo $K$ \textnormal{(in parallel)}}{
      Sample $z_{j,i}\sim E(\cdot\mid\tilde q_j,s_{j,i})$ once, with both adapters disabled\;
      Extract candidate answer $a_{j,i}$ from $z_{j,i}$\;
      Score $r_{j,i}\leftarrow A_\phi(\tilde q_j,z_{j,i})$ using the \texttt{Yes}--\texttt{No} token-logit difference\;
    }
    $i^\star\leftarrow\arg\max_{1\le i\le K}r_{j,i}$\;
    $\hat a_j\leftarrow a_{j,i^\star}$\;
  }
  $\mathcal V_{\mathrm{done}}\leftarrow\mathcal V_{\mathrm{done}}\cup\mathcal R$\;
}
\eIf{no subtasks were specified}{
  $\hat y\leftarrow\hat a_1$\;
}{
  $\hat y\leftarrow\textsc{ComposeTaskAnswer}(q,\{\hat a_j\}_{j=1}^{N})$\;
}
\Return{$\hat y$}\;
\end{algorithm*}